\documentclass[twoside,11pt]{article}
\usepackage{natbib}
\usepackage{graphicx}
\usepackage{color}
\usepackage{rotating}
\usepackage{bbm}
\usepackage{latexsym}
\usepackage{amsmath}               % ACTIVATION EFFECTS LABELS  
\usepackage{amssymb}               % ACTIVATION EFFECTS LABELS  
\usepackage{psfrag} 
\usepackage{wasysym}

\usepackage[ruled,vlined]{algorithm2e}

\usepackage{wrapfig}

\definecolor{brightred}{rgb}{1.0,0.1,0.1}
\definecolor{brightblue}{rgb}{0.0,0.0,0.8}
\definecolor{darkblue}{rgb}{0.0,0.0,0.5}
\definecolor{darkgreen}{rgb}{0.0,0.3,0.0}
\definecolor{brightgreen}{rgb}{0.0,0.8,0.0}
\definecolor{darkblack}{rgb}{0.0,0.0,0.0}
\definecolor{grey}{rgb}{0.3,0.3,0.3}

\graphicspath{{./}}
\DeclareGraphicsExtensions{.eps,.png}
\newcommand{\Ncal}{{\cal N}}

\newcommand{\FF}{{\cal F}}
\newcommand{\FFHat}{\FF}

\newcommand{\FFHatHat}{\FF}

\newcommand{\JBar}{\overline{J}}

\newcommand{\LL}{{\cal L}}

\newcommand{\RRR}{\mathbbm{R}}

\newcommand{\calO}{{\cal O}}

\newcommand{\Zn}{Z^{(n)}}

\newcommand{\eps}{\epsilon}
\newcommand{\epsn}{\eps_n}

\newcommand{\qhatn}{\hat{q}^{(n)}}

\newcommand{\disT}{\textstyle}
\newcommand{\disS}{\displaystyle}

\newcommand{\refp}[1]{(\ref{#1})}
\newcommand{\DKL}[2]{D_{\mathrm{KL}}\big(#1,#2\big)}
\newcommand{\sVec}{\vec{s}}
\newcommand{\sVecN}{\vec{s}^{\,(n)}}
\newcommand{\sVecPrime}{\vec{s}^{\,\prime}}
\newcommand{\sVecT}{\sVecPrime}

\newcommand{\sVecOld}{\sVec^{\,\mathrm{old}}}
\newcommand{\sVecNew}{\sVec^{\,\mathrm{new}}}

\newcommand{\yVec}{\vec{y}}
\newcommand{\yVecN}{\vec{y}^{\,(n)}}

\newcommand{\argmax}[1]{ \underset{#1}{\mathrm{argmax}} }

\newcommand{\E}[1]{\left\langle{}#1\right\rangle}

\newcommand{\bz}{\hspace{-1.2mm}}
\newcommand{\backHalf}{\hspace{-1.5mm}}
\newcommand{\backOne}{\hspace{-3mm}}

\newcommand{\KK}{{\cal K}}
\newcommand{\KKn}{{\KK}^{\hspace{-0.8ex}\phantom{1}^{(n)}}}
\newcommand{\KKnew}{\KK^{\mathrm{new}}}
\newcommand{\KKnewsub}{\KK_{\mathrm{new}}}
\newcommand{\KKoldsub}{\KK_{\mathrm{old}}}
\newcommand{\KKnewN}{\KK^{\hspace{-0.8ex}\phantom{1}^{(n)}}_{\mathrm{new}}}
\newcommand{\KKold}{\KK^{\mathrm{old}}}
\newcommand{\KKtilde}{\tilde{\KK}}

\newcommand{\ThetaOld}{\Theta^{\mathrm{old}}}
\newcommand{\ThetaNew}{\Theta^{\mathrm{new}}}

\newcommand{\ThetaHat}{\hat{\Theta}}
\newcommand{\ThetaHatOld}{\ThetaHat^{\mathrm{old}}}
\newcommand{\ThetaHatNew}{\ThetaHat^{\mathrm{new}}}

\newcommand{\LambdaNew}{\Lambda^{\mathrm{new}}}

\newcommand{\qn}{q^{(n)}}
\newcommand{\pn}{p^{(n)}}

\newcommand{\qt}{\tilde{q}}
\newcommand{\qnt}{\tilde{q}^{(n)}}
\newcommand{\beq}{\begin{equation}}
\newcommand{\eeq}{\end{equation}}
\newcommand{\beqo}{\begin{displaymath}}
\newcommand{\eeqo}{\end{displaymath}}
\newcommand{\bea}{\begin{eqnarray}}
\newcommand{\eea}{\end{eqnarray}} 
\newcommand{\beao}{\begin{eqnarray*}}
\newcommand{\eeao}{\end{eqnarray*}}

\newcommand{\BOX}{$\square$}

\newcommand{\myvanish}[1]{}

\newcommand{\myargmax}[1]{ \underset{#1}{\mathrm{argmax}} }

\begin{document}
%      
%\jmlrheading{-}{-}{}{}{}{J\"org L\"ucke}
%  
%\ShortHeadings{Truncated Variational EM}{L\"ucke}
%\firstpageno{1}    
%        
%        
%\begin{document}
%
%
%
%
\title{Truncated Variational Expectation Maximization\\ \ }
%
%\title{On the Family of Truncated Posteriors\\
%As Variational Approximations\\ \ }
%
%\title{Expectation Truncation -- A Novel Variational EM Approach}
%
%\author{\name J\"org L\"ucke \email joerg.luecke@uni-oldenburg.de\\
%              \addr Machine Learning Group\\
%                    Department of Medical Physics and Acoustics and\\
%                    Cluster of Excellence Hearing4all\\
%                    Carl von Ossietzky Universit\"at Oldenburg\\
%                    26111 Oldenburg\\
%                    Germany
%}
%
%
%
\author{\ \\[3mm]
J\"org L\"ucke\\
         Machine Learning Division\\
%                    Department of Medical Physics and Acoustics and\\
%                    Cluster of Excellence Hearing4all\\
                    Carl von Ossietzky Universit\"at Oldenburg\\
                    26111 Oldenburg, Germany\\[-2mm]
}
%
%
%\editor{}
%
\date{\phantom{nothing}}
\maketitle
\begin{abstract}%
\noindent
We derive a novel variational expectation maximization approach based on truncated variational distributions.
Truncated distributions are proportional to exact posteriors within subsets of a discrete state space and equal zero otherwise.
The treatment of the distributions' subsets as variational parameters distinguishes the approach from previous variational approaches.
The specific structure of truncated distributions allows for deriving novel and mathematically grounded results, which in turn
can be used to formulate novel efficient algorithms to optimize the parameters of probabilistic generative models. Most centrally,
we find the variational lower bounds that correspond to truncated distributions to be given by very concise and efficiently
computable expressions, while update equations for model parameters remain in their standard form. Based on these findings, we show
how efficient and easily applicable meta-algorithms can be formulated that guarantee a monotonic increase of the variational bound. 
Example applications of the here derived framework provide novel theoretical results and learning procedures for latent variable models
as well as mixture models. Furthermore, we show that truncated variation EM naturally interpolates between standard EM with full posteriors
and EM based on the maximum a-posteriori state (MAP). The approach can, therefore, be regarded as a generalization of the popular `hard EM'
approach towards a similarly efficient method which can capture more of the true posterior structure.
\end{abstract}
\section{Introduction}
\label{SecIntro}
The application of expectation maximization \citep[EM;][]{DempsterEtAl1977} is a standard approach to optimize the parameters of probabilistic data models.
The EM meta-algorithm seeks parameters that optimize the data likelihood given the data model and given a set of data points. Data models are typically defined
based on directed acyclic graphs, which describe the data generation process using probabilistic descriptions of sets of hidden and observed variables and their interactions.
EM approaches for most non-trivial such generative data models are intractable, however, and tractable approximations to EM are, therefore, very wide-spread.
EM approximations range from sampling-based approximations of expectation values and related non-parameteric approaches \citep[e.g.][]{GhahramaniJordan1995}, over maximum a-posterior or
`hard EM' approaches \citep[e.g.][]{JuangRabiner1990,CeleuxGovaert1992,OlshausenField1996,LeeEtAl2007,MairalEtAl2010}, Laplace approximations \citep[e.g.][]{KassSteffey1989,FristonEtAl2007}
to variational EM approaches \citep[][and many more]{SaulEtAl1996,NealHinton1998,JordanEtAl1999,OpperWinther2005,Seeger2008,KingmaWelling2014}. %add more recent ones
%
%Variational approaches to expectation maximization \citep[V-EM;][]{} have been developed to improve or enable the development of efficient EM approximations.
%Variational approximations are typically applied to models for which the exact expectation maximization \citep[EM;][DempsterEtAl1977] approach results in
%intractable learning algorithms. The typical reason for intractabilities are large sums or integrals over the set of all hidden states in probabilistic generative models.
%
%As EM for most non-trivial generative models is intractable, approximations to EM are very wide-spread. Alongside variational approaches, EM approximations range from sampling-based 
%approximations of expectation values \citep[][]{} and realted non-parameteric approaches \citep[][]{}, over maximum a-posterior approaches \citep[][]{OlshausenField,Lasso} to Laplace
%approximations \citep[][]{OlshausenField,Lasso,Bishop2006}. 

Instead of aiming at a direct maximization of the data likelihood, variational EM seeks to maximize a lower-bound of the likelihood which is commonly referred to as
the variational {\em free energy} \citep[e.g.][]{NealHinton1998} or the evidence lower bound \citep[{\em ELBO};][]{HoffmanEtAl2013}. Variational lower bounds can be formulated such that their optimization becomes tractable
by avoiding the summation or integration over intractably large hidden state spaces. 
Since the framework of variational free energy approximations has first been explicitly introduced in Machine Learning \citep[e.g.][]{SaulEtAl1996,NealHinton1998,JordanEtAl1999}, 
variational approximations for the EM meta-algorithm have been widely applied and were generalized in many different ways.
Variational EM is now routinely used to train latent variable models (or multiple-causes) models, to train time-series models or to train complex graphical models including models
for deep unsupervised learning \citep[see, e.g][for overviews]{Jaakkola2001,Bishop2006,Murphy2012,KingmaWelling2014,RezendeMohamed2015,PatelEtAl2016}.
Prominent examples of variational EM are based on factored variational distributions \citep[][]{JordanEtAl1999} or Gaussian variational distributions \citep[e.g.][]{OpperWinther2005,Seeger2008,OpperArchambeau2009,KingmaWelling2014}. Truncated distributions were introduced later than factored or Gaussian approaches \citep[][]{LuckeSahani2008,LuckeEggert2010}, and they used instead of a variational loop a sparsity assumption \citep[][]{LuckeSahani2008} or a preselection of latent states \citep[][]{LuckeEggert2010,SheikhEtAl2014,DaiLucke2014,SheltonEtAl2017,SheikhEtAl2019}.

Among approaches which are usually not considered variational are sampling based approximations \citep[e.g.][]{ZhouetAl2009}, or approaches which use just one state (the one with the approximate maximal posterior value) for the optimization of model parameters \citep[][]{JuangRabiner1990,CeleuxGovaert1992,OlshausenField1996,AllahverdyanGalstyan2011,OordEtAl2014}. The latter is commonly referred to as `MAP training' \citep[][]{OlshausenField1996,AllahverdyanGalstyan2011}, `hard EM' \citep[][]{PoonDomingos2011,AllahverdyanGalstyan2011,OordEtAl2014},
`zero temperature EM' \citep[][]{TurnerSahani2011}, as `classification EM' \citep[][]{CeleuxGovaert1992} for mixture models, or as `Viterbi training' for hidden Markov Models \citep[][]{JuangRabiner1990,JordanEtAl1997,CohenSmith2010,AllahverdyanGalstyan2011}. In deep learning, `hard EM' was used, for instance, for generative formulations of convolutional
neural networks \citep[][]{PatelEtAl2016}, deep Gaussian Mixture Models \citep[][]{OordEtAl2014}, or Sum-Product Networks \citep[][]{PoonDomingos2011}.
Note in this respect that also variational approximations (primarily factored ones) have been considered for (deep) graphical models, and deep models have been generalized to 
fully Bayesian settings \citep[e.g.][]{Attias2000,Jaakkola2001,BealGhahramani2003}.
For the purposes of this study, we will introduce a novel variational EM approach assuming a basic generative model of one set of observed and one set of hidden variables (without
distinguishing subsets of these latents). Such a setup may suggest a bipartite graphical models with no further structure,
i.e., a data model in which all hidden variables have the same form of influence on the observed variables \citep[compare, e.g.,][]{NealHinton1998}. Such models are presumably best
suited to follow the introduction of the basic ideas of the novel approach and to highlight its elementary properties. However, as we will not make assumptions about the set of hidden
variables, we here stress that the derived results apply for any directed graphical model with discrete latents, i.e., application to more intricate models including time-series models
or deep directed models are straight-forward. The same does not apply for generalizations to fully Bayesian settings, which would require a major (and potentially very interesting) future research effort. 

In Sec.\,\ref{SecSummary} we provide the problem statement and provide the main result in the form of a meta-algorithm. The reader interested in applying the approach
will find the required information there and a partial and explicit form of the meta-algorithm at the end of Sec.\,\ref{SecOptimization}. In Sec.\,\ref{SecVarApproaches}, we first
discuss standard and recent related versions of variational EM. The section serves (A)~for highlighting the differences and novelties of truncated variational distributions, and 
(B)~will point out where some standard derivations have to be generalized. Sec.\,\ref{SecTVEM} provides results required for a fully variational treatment of truncated distributions.
Sec.\,\ref{SecOptimization} then derives the main theoretical results which are required to formulate the TV-EM meta-algorithm. Sec.\,\ref{SecApplications} presents example applications
of the novel variational framework and shows that `hard EM' is recovered as a special case of TV-EM. We conclude by discussing the results in Sec.\,\ref{SecDiscussion}. 
%
%, examples for applications to different models as well as the application to derive `hard' EM. 
%
%
%
%
\section{Truncated Variational EM -- Summary of the Algorithm}
\label{SecSummary}
%
%\subsection{Problem Description and Notation}
%\label{SecProblem}
%
%In this paper we will provide a number of theoretical results for truncated approximations. 
%
The theoretical results of Secs.\,\ref{SecTVEM} and \ref{SecOptimization} will allow for the formulation of novel variational EM algorithms applicable to generative models with discrete latents, and for identifying existing algorithms as variational approaches. All derivations of these later sections are required to obtain the final results but the final result can be applied without detailed knowledge of the derivations' details. We therefore summarize the main result in the following, and present the required derivations and more details and variants later. 
\subsection{Problem Description and Notation}
%
%First we describe the problem addressed and the notation used.
%
%This section also serves for introducing the notation.
%
Following the framework of Expectation Maximization \citep[EM;][]{DempsterEtAl1977}, our aim is to maximize
the data likelihood defined by a set of $N$ data points, \mbox{$\{\yVec^{\,(1)},\ldots,\yVec^{\,(N)}\}$}, 
where we model the data distribution by a probabilistic generative model whose distribution $p(\yVec\,|\,\Theta)$ is parameterized by the model parameters $\Theta$. We assume generative models
with discrete latent variables $\sVec$ and (discrete or continuous) observed variables $\yVec$ such that the modeled data distribution is given by:
\begin{eqnarray}
  p(\yVec\,|\,\Theta) &=& \sum_{\sVec} p(\yVec,\sVec\,|\,\Theta), %\,=\,\sum_{\sVec} p(\yVec\,|\,\sVec,\Theta),\log(p(\sVec\,|\,\Theta)
\label{EqnGenModle}
\end{eqnarray}
where $\sum_{\sVec}$ notates the sum over all possible values of $\sVec$. The data log-likelihood is then given by:
\begin{eqnarray}
  \LL(\Theta) &=& \log\left(p(\yVec^{(1)},\ldots,\yVec^{(N)}\,|\,\Theta)\right)\ = \ \sum_{n=1}^N\log\left(\sum_{\sVec} p(\yVecN,\sVec\,|\,\Theta)\right).
\label{EqnLikelihood}
\end{eqnarray}
%
%where the data points $\yVecN$ are assumed to be generated independently from a stationary process.
%
%, and were the generative model defines the joint $p(\yVecN,\sVec\,|\,\Theta)$
%
%In the E-step of an EM algorithm
%The likelihood $\LL(\Theta)$ can be optimized using EM \citep[EM;][]{DempsterEtAl1977}. 
%
In its most elementary form, the E-step of an EM algorithm consists of computing the posterior probability $p(\sVec\,|\,\yVecN,\Theta)$
for each data point $\yVecN$ and each latent state $\sVec$, while the M-step updates the parameters $\Theta$. Iterating E- and M-steps monotonically increase the likelihood. 
The basic EM algorithm was generalized in a number of contributions to provide justification for incremental or online versions
and, more importantly for this paper, to provide the theoretical foundation of variational EM approximations \citep[see, e.g.,][]{Hathaway1986,SaulEtAl1996,NealHinton1998}.

\begin{figure}[p]
\begin{center}
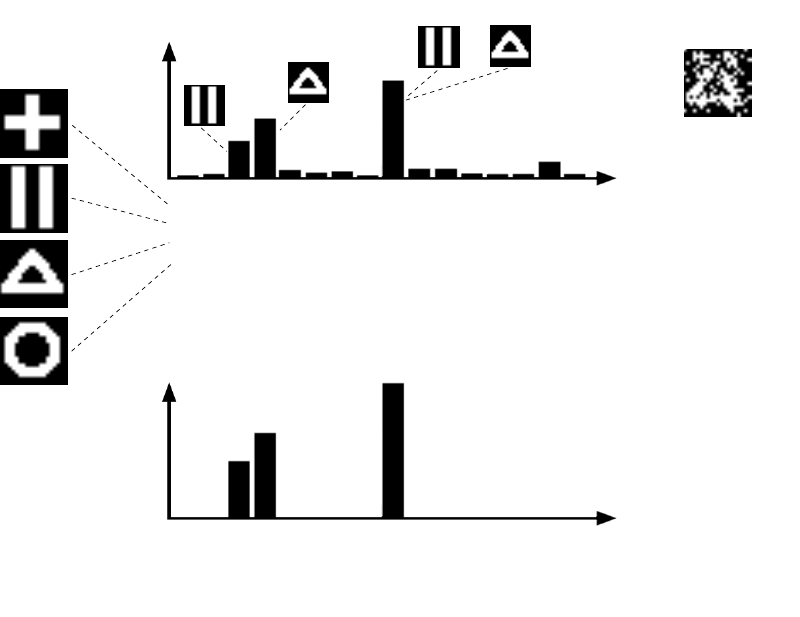
\end{center}
\caption{Illustration of a truncated posterior approximation. Consider data in the form of images where each image can contain any combination of four patterns (`cross', `columns', `triangle', `circle'), see left-hand-side for the patterns. Each combination is denoted by a vector $\sVec$ with binary entries. Given a noisy image, the inference task is to assign to all $\sVec$ a probability value, with each value representing the probability that the image
has been generated by a particular pattern combination. Probabilistic inference can be modeled using a suitable generative model for the data (which we will not further specify here). For the illustration we assume that the right model with the optimal parameters $\Theta$ has already been found. Given a data point $\yVecN$ (upper right), the full posterior $p(\sVec\,|\,\yVecN,\Theta)$ then realizes probabilistic inference using Bayes' rule and the generative model. In the example, high probabilities are assigned to states $\sVec$ containing `columns' and `triangle'. The `cross' pattern is the least consistent with the data point, which results in low probabilities for any state containing a `cross'. A truncated approximation $\qn(\sVec;\KK,\Theta)$ is obtained using any subset ${\KK}^{\hspace{-0ex}^{(n)}}$ of the states. In this example ${\KK}^{\hspace{-0ex}^{(n)}}$ contains the three states with the highest probabilities (lower right), which is the optimal choice for ${\KK}^{\hspace{-0ex}^{(n)}}$ given the constraint $|{\KK}^{\hspace{-0ex}^{(n)}}|=3$. The approximation quality for other truncated approximations strongly depends on the chosen set ${\KK}^{\hspace{-0ex}^{(n)}}$.
%
% give rise to other truncated approximations qualities strongly depending on ${\KK}^{\hspace{-0ex}^{(n)}}$.
%
%Also any other choice of states (any other ${\KK}^{\hspace{-0ex}^{(n)}}$) would give rise to a truncated approximation but with strongly varying approximation qualities.
%but the approximation quality will strongly depend on ${\KK}^{\hspace{-0ex}^{(n)}}$.
}
\label{FigTVEM}
\end{figure}
\subsection{Truncated Variational EM}
The basic idea of truncated EM is the use of truncated posterior distributions as approximations to the full posteriors \citep[e.g.][]{LuckeEggert2010,DaiLucke2014,SheikhEtAl2014,SheltonEtAl2017}.
A truncated approximate posterior distribution is given by:\vspace{-2ex} %The variational distributions are given by:
\begin{eqnarray}
\label{EqnQMain}
\hspace{-4mm}  \qn(\sVec)\hspace{-2mm} &:=& \hspace{-2mm}\qn(\sVec;\KK,\Theta)=\frac{\disS\phantom{\int}p(\sVec\,|\,\yVecN,\Theta)\phantom{\int}}
{\hspace{-2mm}\disS\sum_{\sVecT\in\KKn}p(\sVecT\,|\,\yVecN,\Theta)}\,\delta(\sVec\in\KKn)=\frac{\disS\phantom{\int}p(\sVec,\yVecN\,|\,\Theta)\phantom{\int}}
{\hspace{-2mm}\disS\sum_{\sVecT\in\KKn}p(\sVecT,\yVecN\,|\,\Theta)}\,\delta(\sVec\in\KKn),
\end{eqnarray}
where $\delta(\sVec\in\KKn)$ is an indicator function, i.e., $\delta(\sVec\in\KKn)=1$ if $\sVec\in\KKn$ and zero otherwise.
Fig.\,\ref{FigTVEM} shows an illustration of a truncated posterior approximation.
The set $\KKn$ contains a finite number of hidden states $\sVec$. There is one such set for each data point $n$, and we will
denote with $\KK$ the collection of all sets $\KKn$, i.e., $\KK=\KK^{(1:N)}=(\KK^{(1)},\ldots,\KK^{(N)})$ (we will use the `colon' notation
to denote a range of indices throughout this paper). The expectation values w.r.t.\,truncated distributions are given by:
\begin{equation}
\E{g(\sVec)}_{\qn}\,=\,\E{g(\sVec)}_{\qn(\sVec;\,\KK,\Theta)}\,=\,\frac{\disS\sum_{\sVec\in\KKn} p(\sVec,\yVecN\,|\,\Theta)\ g(\sVec)
}{\disS\sum_{\sVecT\in\KKn}p(\sVecT,\yVecN\,|\,\Theta)}\,,
\label{EqnSuffStatMain}
\end{equation}
where $g(\sVec)$ can be any (well-behaved) function over latents $\sVec$. For sufficiently small sets $\KKn$, the expectation values (\ref{EqnSuffStatMain}) are computationally tractable if the joint distribution $p(\sVec,\yVecN\,|\,\Theta)$ of a probabilistic generative model is efficiently computable. As for most directed graphical models the joint is indeed computational tractable, we will assume such tractability for the paper (unless stated otherwise).
%expectation values (\ref{EqnSuffStatMain}) are computationally tractable if $\KKn$ contains a sufficiently low number of states.

The variational distributions (\ref{EqnQMain}) define, similar to other types of variational distributions, a variational lower bound, which is given by:
\begin{eqnarray}
\label{EqnFreeEnergyTVEMOrgA}
  \FF(\KK,\ThetaOld,\Theta)\,=\,
    \sum_{n=1}^{N} \bigg[
      \sum_{\sVec}\ \qn(\sVec;\KK,\ThetaOld)\  
       \log\left( p(\sVec,\yVecN\,|\,\Theta)\right) 
      \bigg] 
      + H(q(\sVec;\KK,\ThetaOld))\,,
\end{eqnarray}
where $H(q)$ is an entropy term in which $\Theta$ is held fixed at $\ThetaOld$. The bound will be referred as {\em free energy} from now on.

For $\ThetaOld=\Theta$ the free energy can be shown (see Sec.\,\ref{SecOptimization}, Prop.\,3) to take on a simplified form given by:
\begin{eqnarray}
\label{EqnFreeEnergyTVEMOrgB}
\FF(\KK,\Theta) = \FF(\KK,\Theta,\Theta) &=& \disS\sum_{n=1}^{N}\log\big(\backHalf\sum_{\ \sVec\in\KKn} p(\sVec,\yVecN\,|\,\Theta)\,\big)\,.%
\end{eqnarray}
The free energy (\ref{EqnFreeEnergyTVEMOrgB}) we will refer to as {\em simplified} truncated free energy or just truncated free energy.
The truncated free energy lower bounds the log-likelihood (\ref{EqnLikelihood}) and it is provably monotonically increased by the following procedure:
%
%The main result of this paper is that the following procedure monotonically increases the 
%
%In order to derive a learning algorithm for GMMs based on truncated distributions, we have to answer the question how the parameters $\KKn$ and $\Theta$ are to be updated.
%For our purposes we will here make use of results in \citep[][]{Lucke2016}, which address this question for general generative models (with discrete latents). More specifically we
%will use the result that the following truncated variational EM step (TV-EM step) is guarranteed to monotoneously increase a lower-bound (the free energy $\FF$) of the data log-likelihood:
%
\begin{align}
	& \KKnew = \myargmax{\KK}\,\big\{\FF(\KK,\ThetaOld)\big\}  && \hspace{0mm}\text{TV-E-step} \label{EqnTVEStep}\\
%\FF(\KK,\Theta) = \disT\sum_{n=1}^{N}\log\big(\sum_{c\in\KKn} p(c,\yVecN\,|\,\Theta)\big). \label{EqnFTrunc}\\
%
  &\ThetaNew = \myargmax{\Theta}\,\big\{\FF(\KKnew,\ThetaOld,\Theta)\big\} && \hspace{0mm}\text{TV-M-step} \label{EqnTVMStep}\\
 &\ThetaOld = \ThetaNew && \hspace{0mm}\text{}\label{EqnTVThird}
\end{align}
We will refer to one iteration of Eqns.\,\ref{EqnTVEStep} to \ref{EqnTVThird} as a {\em truncated variational EM} (TV-EM) iteration. The repetition of TV-EM iterations until convergence of $\Theta$ 
monotonically increases the lower free energy bound (\ref{EqnFreeEnergyTVEMOrgB}) of the likelihood to at least local optima.

The optimization of $\FF(\KK,\Theta)$ w.r.t.\,$\KK$ has hereby to be taken as optimization for sets $\KKn$ with limited size. Small $\KKn$ of constrained size will ensure computational tractability as well as non-trivial solutions of (\ref{EqnTVEStep}). The $\KKn$ can hereby be thought of as being all constraint to the same constant size, $|\KKn|=const$ for all $n$, although the results which we will derive 
in this study will also allow for other size constraints.
%
%be independent of any specific form of size constraints.
%
For sufficiently small state sets $\KKn$, the TV-E-step (\ref{EqnTVEStep}) is a constraint discrete optimization of a computationally tractable function (given by Eqn.\,\ref{EqnFreeEnergyTVEMOrgB}).
Furthermore, regarding the TV-M-step, any closed-form or gradient updates of $\Theta$ derived using (\ref{EqnTVMStep}) are in this case computationally tractable because the expectation values w.r.t.\ $\qn(\sVec;\KK,\Theta)$ are tractable according to Eqn.\,\ref{EqnSuffStatMain}. For many standard generative models (e.g., sparse coding models, mixture models, hidden Markov models etc.) the M-steps are well-known and often derivable in closed-form.
% \citep[][etc]{GMM,pPCA,sparseCoding}.
The TV-M-step (\ref{EqnTVMStep}) warrants that such M-step equations remain unchanged if TV-EM is applied -- only the expectation values (which the M-steps depend on) have to be replaced by Eqn.\,\ref{EqnSuffStatMain}.
Like for other variational EM approximations, the variational E- and M-steps (Eqns.\,\ref{EqnTVEStep} and \ref{EqnTVMStep}), can also be changed to just partially optimize the free energy, i.e., to increase instead of maximize the free energy.
For partial TV-E- and/or TV-M-steps the guarantee that the free energy monotonically increases does continue to hold (see Eqns.\,\ref{EqnTVEMOptStepsParialFinal}).

Eqns.\,\ref{EqnSuffStatMain} to \ref{EqnTVThird} sufficiently summarize the TV-EM meta-algorithm such that it can directly be applied to a generative model with discrete latents. Sec.\,\ref{SecExplicitForm} reiterates the algorithm in a very explicit form which highlights that the algorithm is formulated solely in terms of the joint probabilities $p(\yVec,\sVec\,|\,\Theta)$ given by the considered generative model. In Sec.\,\ref{SecApplications} we will discuss different realizations of concrete TV-EM algorithms. The reader interested in applying TV-EM may directly be referred to these sections.

We will now proceed and derive TV-EM (Eqns.\,\ref{EqnTVEStep} to \ref{EqnTVThird} with Eqn.\,\ref{EqnFreeEnergyTVEMOrgB}) step by step. All of the following derivations will be necessary to prove the properties of the TV-EM algorithm, and none of the Eqns.\,\ref{EqnFreeEnergyTVEMOrgB} to \ref{EqnTVThird} will turn out to be trivial. This includes Eqns.\,\ref{EqnTVMStep} and \ref{EqnTVThird} although they may seem straight-forward at first sight.
%
%
%
%
%

%with each iteration.
%without change to the guarantee that the free energy monotonically increases with each iteration.
%
%
%
\section{Variational Approaches to Expectation Maximization and Related Work}
\label{SecVarApproaches}
We briefly review well-known as well as some recent results about variational EM.
%
% The section will facilitate the introduction of truncated variational approaches,
%and will work out its distinguishing features.
%
%
%
%\subsection{The Variational Free Energy Optimization}
%\label{SecVEM}
%
Following the introduction of the EM algorithm \citep[e.g.][]{DempsterEtAl1977}, the central novel object introduced by variational approaches was the 
variational {\em free energy} (e.g., \citealt[][]{SaulEtAl1996,NealHinton1998}, also compare \citealt[][]{Hathaway1986}). The free energy is a lower
bound of the log-likelihood, and provides the theoretical foundation of variational EM approximations. The crucial result relating free energy to the
data likelihood \refp{EqnLikelihood} is (in our notation) given by:
\begin{eqnarray}
\label{EqnDKLStandard}
\LL(\Theta)-\FF(q,\Theta)\ =\ \sum_{n}\DKL{\qn(\sVec)}{p(\sVec\,|\,\yVecN,\Theta)}\ \geq\ 0,
\end{eqnarray}
where $\DKL{q}{p}$ denotes the Kullback-Leibler (KL) divergence. 
Importantly for this work, note that the standard derivation of \refp{EqnDKLStandard} requires strictly positive variational distributions, i.e., \mbox{$\qn(\sVec)>0$} for all $\sVec$ and $n$
(we show the standard derivation in \ref{AppA}). Later on, in Sec.\,\ref{SecTVEM}, we will show how this restriction can be lifted.  
Using \refp{EqnDKLStandard} we observe that the free energy $\FF(q,\Theta)$ is maximized
if we choose (for all $n$) the posterior distribution as variational distribution, $\qn(\sVec) = p(\sVec\,|\,\yVecN,\Theta)$ which is the subject, e.g., of Lemma~1 by \citet{NealHinton1998}.
In the case of generative models with tractable posterior, the free energy $\FF(q,\Theta)$ is used to 
maximize the likelihood \refp{EqnLikelihood} by iteratively optimizing this objective w.r.t.\ paratmeters $\Theta$, and by setting $q$ to the exact posteriors.
In comparison to the optimization of data likelihoods, the optimization of free energies is usually easier because their derivatives can be taken while the parameters $\Theta$
of the posterior distribution are held fixed. %This results, e.g., in the well known closed-form M-steps for Gaussain Mixture models and GMMs, hidden Markov models (HMMs) or probabilistic PCA.
%After maximizing $\FF(q,\Theta)$ w.r.t.\ $\Theta$ (the M-step), we can set $\qn(\sVec)=p(\sVec\,|\,\yVecN,\ThetaOld)$ with $\ThetaOld=\Theta$ because this choice maximizes
%the free energy according to (\ref{EqnFreeEnergyExactEM}). 

Variational approximations are motivated by generative models whose posterior probabilities are not tractable. To still efficiently optimize the model parameters, 
the basic idea of variational EM is to maximize the free energy $\FF(q,\Theta)$ in \refp{EqnFreeEnergy} for a constrained class of variation distributions~$q$.
If we can find variational distributions $q$ such that (A)~optimization of $\FF(q,\Theta)$ is tractable, and (B)~such that the lower bound $\FF(q,\Theta)$ becomes
as similar (as tight) as possible to $\LL(\Theta)$, then a tractable approximate optimization of $\LL(\Theta)$ is obtained.

By definition of the free energy, almost no restrictions are imposed on the choice of the distributions $q$, such that they can,
in principle, be chosen to take any functional form and to be dependent on any set of parameters.
To fulfill requirement~(A) some choice for a functional form of $q$ has to be made, however. %the variational distributions are typically restricted to specific functional forms of distribution.
As a consequence, the distributions $\qn(\sVec)$ become equipped with additional parameters and these parameters
are then optimized to make $\FF(q,\Theta)$ as tight as possible (requirement B). The variational distributions are subsequently
denoted by $\qn(\sVec,\Lambda)$ with the additional parameters $\Lambda$ being referred to as {\em variational parameters}.
Having chosen a variational distribution, the free energy is often taken to depend on $\Lambda$ rather than on the
variational distributions themselves, i.e., $\FF(\Lambda,\Theta)$.
%
%Without being more specific about the choice of variational distributions and parameters, 
The iterative procedure to optimize the free energy may then be denoted by:
\begin{align*}
\label{EqnFreeEnergyVEM}
\mathrm{Opt\ 1:} & & & \LambdaNew = \argmax{\Lambda}\big\{\FF(\Lambda,\ThetaOld)\big\} & & \mbox{while holding $\ThetaOld$ fixed} & \mbox{(V-E-step)}\\
\mathrm{Opt\ 2:} & & & \ThetaNew  =\argmax{\Theta}\big\{\FF(\LambdaNew,\Theta)\big\} & & \mbox{while holding $\LambdaNew$ fixed} & \mbox{(V-M-step)} \\
\mathrm{} & & & \ThetaOld = \ThetaNew
\end{align*}
The two optimization steps are repeated until the parameters $\Theta$ have sufficiently converged.

%The second optimization step (the M-step) is a standard optimization of model parameters
%while the first optimization (the variational E-step or V-E-step) is an additional procedure that updates
%the variational parameters. One way to optimize
%$\FF(\Lambda,\Theta)$ is to vary $\Lambda$ and to choose the $\Lambda$ corresponding to the largest
%$\FF(\Lambda,\Theta)$ under such a variation. This procedure
%can often be facilitated based on analytical results. % \citep[][]{}.
%Full and partial optimization of $\FF(\Lambda,\Theta)$ both increase the free energy continuously.
%
In principle, many possible choices of variational distributions that fulfill requirements (A)
and (B) are conceivable; and any choice would give rise to a variational EM procedure.
Still, the application of variational EM has been dominated by two standard types of variational distributions:
Gaussian variational distributions and factored variational distributions.% ({\em Factored Variational EM}).
%
%The optimization problem for $\Lambda$ takes
%the form of a classical variational optimization problem (look at Paisley for CITE) -- which
%motivated the name ``variational expectation maximization'' or ``variational EM'' \citep[see][]{NealHinton1998}.
%
%
%
\subsection{Gaussian and Factored Variational EM}
As one of the most basic distributions is the Gaussian distribution, a natural choice for variational distributions
(for continuous latents) is a multi-variate Gaussian: 
\begin{eqnarray}
\label{EqnFreeEnergyGVEM}
  \qn(\sVec) &:=& \qn(\sVec;\Lambda)\ =\ \qn(\sVec;\mu^{(n)},\Sigma^{(n)})\ =\ \Ncal(\sVec;\mu^{(n)},\Sigma^{(n)})\,,
\end{eqnarray}
where $\Lambda^{(n)}=(\mu^{(n)},\Sigma^{(n)})$ and where $\Lambda=(\Lambda^{(1)},\ldots,\Lambda^{(N)})$ is the set of
all variational parameters (one mean and one covariance matrix per data point). The Gaussian variational approach
approximates each posterior distribution by a Gaussian distribution. %(essentially capturing its first two moments).
The optimization of the variational parameters in
(\ref{EqnFreeEnergyGVEM}) results in update equations for mean and variance that maximize the free energy and minimize
the KL-divergence between true posteriors and the variational Gaussians. Gaussian distributions are especially well suited
for data models with mono-modal posteriors, and are consequently popular, e.g., for optimization of sparse linear
models \citep[e.g.][]{OpperWinther2005,Seeger2008,OpperArchambeau2009}. 
%Gaussians can capture data correlations and jointly optimize mean and variance in the KL-divergence sense.
More recently, Gaussian variational distributions are the standard and most popular choice to train 
variational autoencoders \citep[][and many more]{KingmaWelling2014}. In these approaches, standard neural networks are used
to parameterize mean and covariance matrix of the Gaussian. Parameters of the networks are then optimized within
the variational E-step.

%Note that the latter is different from Laplace approximations which first fit the mean and then the covariance matrix \citep[see, e.g.,][]{Bishop2006} .
%
% Discuss Seeger (GLM) here, discuss Opper paper, discuss ..., Neil Laurence
%
%More details on Gaussian Variational EM may be found, e.g., in work by ...
%
%

%
The second standard variant of variational EM builds up on the choice
of variational distributions that factor over sets of hidden variables.
Most commonly the choice is a fully factored distribution:
\begin{eqnarray}
\label{EqnFreeEnergyFVEM}
  \qn(\sVec) &:=& \qn(\sVec;\Lambda)\ =\ \prod_{h=1}^{H}q_h(s_h;\vec{\lambda}_h^{(n)})
\end{eqnarray}
where $\vec{\lambda}_h^{(n)}$ are parameters associated with one hidden
variable $h$ for one data point $n$, and where $\Lambda$ is the collection of all these parameters (all $h$ and $n$ combinations).
Using factored distributions in combination with specific choices for the factors
$q_h(s_h;\vec{\lambda}_h^{(n)})$ then results in computationally
tractable optimizations. For the usual generative models the factors are often chosen to be identical
to the prior distributions of the individual hidden variables \citep[e.g.][]{JordanEtAl1999,HaftEtAl2004}.
Because of a mathematical analogy to variational free energy approximations in statistical physics (which initially motivated variational EM),
fully factored variational approaches (\ref{EqnFreeEnergyFVEM}) are also frequently termed {\em mean field} approximations.

\subsection{Further Variational Optimization Approaches}
The assumption of Gaussian or mean-field variational approaches may not match the true posteriors well. Generalizations of these
standard approaches have therefore been investigated. Generalizations of fully factored mean-field approaches can, for instance,
be defined by allowing for dependencies between small sets of variables (doubles, triples etc) resulting in {\em partially factored} approaches.
%or {\em structured variational} approaches.
%
Such approaches can capture more complex posterior interdependencies (correlations as well as higher-order dependencies), and they are therefore also
termed {\em structured variational} or {\em structured mean-field} approaches to highlight their close relation to fully factored approaches \citep[compare][]{SaulJordan1996,MacKay2003,Bouchard2009,Murphy2012}. 

As factorized variational distributions (including structured ones) make potentially harmful assumptions \citep[][]{IlinValpola2003,MacKay2003,TurnerSahani2011,SheikhEtAl2014}, alternative
distributions have repeatedly been investigated. Recent examples are `normalizing flow' approaches for continuous latents \citep[][]{RezendeMohamed2015}, copula-based approaches \citep[][]{TranEtAl2015},
or approaches that hierarchically expand mean-field approaches to include dependencies \citep[][]{RanganathEtAl2015}.  %\comment{Is black-box VI sufficiently discussed?}
Such approaches use specific transformations
of distributions to allow for modeling complex dependencies among latent variables for improved posterior approximations. We will briefly discuss the relation of these approaches to
truncated variational EM in the context of `black box' optimization in Sec.\,\ref{SecDiscussion}. Further work which generated recent attention \citep[][]{HernandezEtAl2016,RanganathEtAl2016a}
considers generalizations of the original likelihood and free energy objectives (Eqns.\,\ref{EqnLikelihood} and \ref{EqnFreeEnergyTVEMOrgA}). Also related in this context is work on variational
approaches using stochastic variational inference \citep[][]{HoffmanEtAl2013}, where auxiliary distributions for Markov chains are defined and used to approximate true posteriors. 
Generalizations of free energy objectives and stochastic variational inference can both be considered complementary lines of research to the results discussed in this work. 
%
%
%
%
%ADD DISCUSSION OF RECENT APPROACHES
%
%Factored approaches and especially the mean-field approach have been, by far, the most popular variational approaches. This is
%also reflected by them being often equated with ``variational EM'' although a specific factorizing choice was made.
%
%\citep[][]{TurnerSahani2011,HoffmanEtAl2013} although a specific factorizing choice was made.
%which to some extend ignores the specific factorizing choice
%made for the variational distributions.
%
%
%
%
%
%
%
\section{Truncated Variational Distributions}
\label{SecTVEM}
The introduction of Gaussian and factored variational distributions now provides the ground for the introduction
of a novel class of variational distributions. In contrast to the prominent examples of variational EM, we will here neither
assume monomodal variational distributions (like Gaussian variational EM) nor independent factors (like mean-field approaches).
Furthermore, we will not choose a specific analytic function such as Gaussians or products of elementary distributions.
Instead, we use the posterior distribution itself to define variational distributions. More precisely, we define the variational
distributions to be proportional to the full posteriors. However, for any given data point $\yVecN$, the proportionality will be
constrained to a subspace $\KKn$ which will allow for computationally tractable procedures. States $\sVec$ not in $\KKn$ are assumed
to have zero probability (`hard' zeros). Formally, such truncated distributions $\qn(\sVec)=\qn(\sVec;\KK,\Theta)$ are given by
Eqn.\,\ref{EqnQMain}. Fig.\,\ref{FigTVEM} provides an illustration.
%
%A consequence of this definition is the dependency of $\qn(\sVec;\KK,\Theta)$ on {\em two} types of variational parameters, the states $\KKn$ and the parameters $\Theta$.
%The latter are of the same form as the model paramters but do not necessarily have to represent the same values. We will, therefore, from now on refer to the variational
%paramters $\Theta$ of the truncated distributions as $\ThetaHat$ to distinguish them from the parameters $\Theta$ of the generative model.
%
%  (but may have different values. 
%
%Furthermore, note that truncated distribu
%in contrast to Gaussian or mean-field approaches, truncated distributions
%

Truncated distributions have been suggested previously and have successfully been applied to a number of elementary as well as
more complex generative data models \citep[][]{LuckeEggert2010,DaiEtAl2013,DaiLucke2014,SheikhEtAl2014,SheikhLucke2016,SheltonEtAl2017}.
Instead of using a variational optimization of approximation parameters similar to Gaussian or mean-field approaches, truncated approximations
have, so far, used preselection mechanisms to reduce the number of states evaluated for a truncated approximation \citep[][]{LuckeEggert2010}.
While truncated EM was shown to be very efficient in practice \citep[][]{SheikhLucke2016,HughesSudderth2016,SheltonEtAl2017}, 
no fully variational treatment was provided, and no convergence guarantees and free energy results as
they are derived here, were given. We will later see, however, that preselection based truncated EM can be 
closely related to the fully variational framework developed in this work.
%
%it was unclear how such an optimization could be defined (we will address this central problem in Sec.\,\ref{SecOptimization}). 
%
%The use of the posterior itself to define an approximation is new for variational approaches. It is worth pointing out, however,
%that in the domain of stochastic approximations, Gibbs sampling \citep[][]{GemanGeman1984} also represents an example where the posterior itself
%is (via its conditionals) used to define an approximation. %Also for Gibbs sampling no explicit functional form of
%a proposal distribution has to be chosen.

%Note that by considering Eqn.\,\ref{EqnQMain}, it can instantly be observed that $\qn(\sVec)$ becomes equal to the exact
%posterior $p(\sVec\,|\,\yVecN,\Theta)$ if $\KKn$ contains all possible states of $\sVec$.
%Furthermore, $\qn(\sVec)$ is equal to the posterior if for all states outside
%$\KKn$ the posterior mass is zero.
%In general, computing all latent states is computationally prohibitive such that $\KKn$ has
%to be a sufficiently small subset, and $\qn(\sVec)$ becomes an approximate posterior. The larger the posterior mass that can be captured
%by the states in $\KKn$, the better is the approximation. Consequently, if the posterior mass is concentrated
%in small volumes of the state space of $\sVec$, approximation (\ref{EqnQMain}) can become very accurate for small $\KKn$.
%In that case, also the computation of expectation values w.r.t.\ truncated distributions becomes tractable, see Eqn.\,\ref{EqnSuffStatMain}.

Having defined the variational distribution of (\ref{EqnQMain}), we can now seek to derive the corresponding free energy.
However, by considering the standard derivation (\ref{EqnFreeEnergyDeri}) given in \ref{AppA}, recall that we required the values of $\qn(\sVec)$ to be {\em strictly} positive,
$\qn(\sVec)>0$ for all $\sVec$. In order to embed truncated distributions (\ref{EqnQMain}) into the free energy framework, we therefore first have to formally generalize the
free energy formalism. % to be applicable without the constraint of strict positivity.
Here we summarize the main results in the form of two propositions. Prop.\,1 shows that the application of Jensen's inequality to obtain a free energy can be generalized
to distributions with exact zeros:
\ \\[-1mm]
\ \\
{\bf Proposition 1}\\*
Let $\qn(\sVec)$ be variational distributions defined on a set of states $\Omega$ (with values $\qn(\sVec)$ not necessarily greater zero),
then a corresponding free energy $\FF(q,\Theta)$ exists and is given by:
\begin{eqnarray}
\label{EqnPropOne}
%
%\normalsize
\FF(q,\Theta)\,:=\,
    \sum_{n=1}^{N} \Big(
      \sum_{\sVec}\ \qn(\sVec)\  
       \log\big( p(\sVec,\yVecN\,|\,\Theta)\big)
      \Big) 
      + H(q)\,.
\end{eqnarray}
\ \\[-7mm]
\BOX\\[3mm]
The set $\Omega$ simply denotes the set of all possible values that $\sVec$ can take on. The formal proof of Prop.\,1 (see \ref{AppB}) uses auxilary distributions $\qt(\sVec)$ with $\qt(\sVec)=\eps$ for all $\sVec\in\Omega$ with $q(\sVec)=0$. The free energy for $q$ is then obtained in the limit $\eps\rightarrow{}0$. As the free energy (\ref{EqnPropOne}) is obtained as a limit point,
it remains to be formally shown that the lower bound property not only applies for any finite $\eps$ but that also the limit is a lower bound of $\LL(\Theta)$. % also in this limit. 

\ \\[-1mm]
{\bf Proposition 2}\\*
Let $\qn(\sVec)$ be variational distributions over a set of states $\Omega$ (with values $\qn(\sVec)$ not necessarily greater zero),
then the corresponding free energy (\ref{EqnPropOne}) is a lower bound of the likelihood, and the difference between likelihood and free energy is given by the sum
over KL-divergences:
\begin{eqnarray}
\label{EqnPropTwo}
\LL(\Theta)-\FF(q,\Theta)\ =\ \sum_{n=1}^{N}\DKL{\qn(\sVec)}{p(\sVec\,|\,\yVecN,\Theta)}\ \geq\ 0.\vspace{-4mm}
\end{eqnarray}
\ \\[-6mm]
\BOX\\[4mm]
The formal proof of Prop.\,2 is given in \ref{AppB}. The proofs of both propositions have (to the knowledge of the author) not been provided before.

The free energy of Prop.\,1 as well as its relation to the likelihood stated by Prop.\,2 take on the same form as those for variational distributions with only positive values.
As such Props.\,1 and 2 are not surprising. As `hard' zeros are a distinguishing feature of truncated variational distributions, Props.\,1 and 2 are, however, strictly required before
any further rigorous results can be obtained. Furthermore, note that both propositions apply for any variational distribution with `hard' zeros (for discrete latents), such that
Props.\,1 and 2 remove the constraint to strictly positive distributions also in general.
%
%
%
%
%Taken together, Propositions 1 and 2 mean that we can generalize the standard free energy framework to any variational distribution --
%the requirement of strictly positive distributions can be dropped. As a consequence,
%we can use the (generalized) free energy framework also for the truncated variational distributions in (\ref{EqnQMain}).
%The variational free energy associated
%with a truncated distribution will be referred to as {\em truncated variational free energy} or simply {\em truncated free energy}.

Based on Props.\,1 and 2, we can now insert the specific functional form of truncated distributions (\ref{EqnQMain}) into the free energy (\ref{EqnPropOne}).
%\comment{HERE SHOULD BE A TEXT STATING WHY THE FOLLOWING IS STRICTLY NECESSARY}\\
As for non-variational EM, i.e., EM with exact posterior as variational distributions, $\qn(\sVec)\,=\,p(\sVec\,|\,\yVecN,\ThetaHat)$, we will distinguish
between parameters $\ThetaHat$ of the variational distribution and the parameters $\Theta$ of the generative data model. Like for the M-step of exact EM,
this allows for taking derivatives of the log-joint $\log(p(\sVec,\yVecN\,|\,\Theta))$ while the variational distributions can be treated as constant
(we will come back to this point further below). Inserting the truncated distributions $\qn(\sVec;\KK,\ThetaHat)$ into the free energy (\ref{EqnPropOne}) then yields:
\begin{eqnarray}
\label{EqnFreeEnergyTVEMOrg}
  \FFHat(\KK,\ThetaHat,\Theta)\,=\,
    \sum_{n=1}^{N} \bigg[
      \sum_{\sVec}\ \qn(\sVec;\KK,\ThetaHat)\  
       \log\left( p(\sVec,\yVecN\,|\,\Theta)\right) 
      \bigg] 
      + H(q(\sVec;\KK,\ThetaHat))\,.%CHANGE TO \ThetaOld?
\end{eqnarray}
%\end{multline}
%
The free energy now depends on {\em three} sets of parameters, $\KK$, $\ThetaHat$ and $\Theta$. %The parameters $\ThetaHat$ and $\Theta$ are of the same type but do not necessarily represent the same values.
The Shannon-entropy $H(q(\sVec;\KK,\ThetaHat))$ is independent of the parameters $\Theta$.
%

%The variational free energy $\FFHat$ now depends on three sets of parameters: two sets of variational parameter $\KK$ and $\ThetaHat$, and the model parameters $\Theta$.
%The parameters $\KK$ are a set of latent states $\sVec$. The parameters $\ThetaHat$ and $\Theta$ are the same parameters but can potentially different values. The
%Shannon-entropy $H(q(\sVec;\KK,\ThetaHat))$ is independent of the model parameters $\Theta$.

\section{Optimization of Truncated Variational Free Energies}
\label{SecOptimization}
%
%\comment{OR HERE SHOULD BE A TEXT STATING WHY THE FOLLOWING IS STRICTLY NECESSARY}\\
The variational parameters of $\KK$ and $\ThetaHat$ that we aim at optimizing are 
different from the typical variational parameters, e.g., different from those of factored
variational approaches or of Gaussian approximations. For each $n$, the set $\KKn$ contains discrete points in latent
space; and the parameters $\ThetaHat$ are of the same type as those of
the generative model but with potentially different values. 

At this point it may be tempting to simply treat the parameters $\ThetaHat$ as the parameters of a full posterior, i.e., to hold fixed at the
current parameters when we update $\Theta$ in the M-step. However, the optimization of $\ThetaHat$ in a variational E-step can potentially result
in values different from $\Theta$. Only for full posteriors it has previously been shown that $\ThetaHat=\Theta$ maximizes the
free energy \citep[][]{NealHinton1998}. For general variational distribuitons that depend on the posterior this is {\em not} the case.
We therefore have to treat (at least for now) $\ThetaHat$ as different from $\Theta$.
%
%If the variational
%distributions depend on the posterior, then setting $\ThetaHat=\ThetaOld$ is wrong in general (i.e., this can results in the variational lower bound to {\em decrease}).
%A counter-example can be provided, for instance, by using annealed posteriors as variational distributions. As a consequence, whenever we want to optimize a variational
%distribution defined using the posterior itself, we have to {\em proof} that setting $\ThetaHat=\ThetaOld$ is optimal. Such a proof is in general not possible (see counter example).
%However, for truncated distributions it is possible to show this. This paper is doing just this: using the derivatives presented it provides the proof (with full posteriors and
%MAP as boundary cases).
%
%
%As the (generalized) variational
%framework of Sec.\,\ref{SecTVEM} does not require strict positivity from the variational distributions, the
%truncated distributions $\qn(\sVec;\KK,\ThetaHat)$ can now be treated within a variational free energy framework.
%
%, which does not impose specifics on the
%distributions chosen as variational distributions.
%
Following the free energy approach we aim at optimizing the free energy (\ref{EqnFreeEnergyTVEMOrg}) instead of directly optimizing the likelihood.
We will do so by optimizing $\FFHat(\KK,\ThetaHat,\Theta)$ step-by-step (coordinate wise) w.r.t.\ its three sets of parameters:\vspace{2mm}
%\begin{equation}
%begin{array}{lcllll}
%\label{EqnTVEMOptSteps}
%%
%\mathrm{Opt\ 1:} & &  \ThetaNew  &=\argmax{\Theta}\FFHat(\KK,\ThetaHat,\Theta) & & \mbox{holding $\KK$ and $\ThetaHat$ fixed}\hspace{-6mm}\phantom{\int^f_g}\\
%%
%\mathrm{Opt\ 2:} & &    \ThetaHatNew &= \argmax{\ThetaHat}\FFHat(\KK,\ThetaHat,\ThetaNew) & & \mbox{holding $\KK$ and $\ThetaNew$ fixed}\hspace{-6mm}\phantom{\int^f_g}\\
%%
%\mathrm{Opt\ 3:} & &    \KKnew &= \argmax{\KK}\FFHat(\KK,\ThetaHatNew,\ThetaNew) & & \mbox{holding $\ThetaHatNew$ and $\ThetaNew$ fixed}\hspace{-6mm}\phantom{\int^f_g}
%%
%\end{array}
%\end{equation}
%
\begin{equation}
\begin{array}{lcllll}
\label{EqnTVEMOptSteps}
\mathrm{Opt\ 1:} & &    \KKnew &= \hspace{1mm}\argmax{\KK}\big\{\FFHat(\KK,\ThetaHatOld,\ThetaOld)\big\} & & \mbox{while holding $\ThetaHatOld$ and $\ThetaOld$ fixed}\hspace{-6mm}\phantom{\int^f_g}\\[2mm]
\mathrm{Opt\ 2:} & &  \ThetaNew  &=\hspace{1mm}\argmax{\Theta}\big\{\FFHat(\KKnew,\ThetaHatOld,\Theta)\big\} & & \mbox{while holding $\KKnew$ and $\ThetaHatOld$ fixed}\hspace{-6mm}\phantom{\int^f_g}\\[2mm]
\mathrm{Opt\ 3:} & &    \ThetaHatNew &= \hspace{1mm}\argmax{\ThetaHat}\big\{\FFHat(\KKnew,\ThetaHat,\ThetaNew)\big\} & & \mbox{while holding $\KKnew$ and $\ThetaNew$ fixed}\hspace{-6mm}\phantom{\int^f_g}\\[6mm]
\mathrm{\ \ \ \ \ \,set} & &          \ThetaHatOld &\hspace{0mm}=\hspace{2mm} \ThetaHatNew \mbox{\ and\ } \ThetaOld\hspace{2mm}=\hspace{2mm}\ThetaNew & &\hspace{-9mm}\mbox{and start-over with Opt 1}
\end{array}
\end{equation}
The order of the updates is chosen for later convenience. Each of the three optimization steps by definition increases the free energy $\FFHat(\KK,\ThetaHat,\Theta)$ w.r.t.\ one of its arguments. Opt~2
which updates the model parameters $\Theta$ corresponds to the M-step. Opt~1 and Opt~3 optimize the two sets of variational parameters $\KK$ and $\ThetaHat$, respectively, and correspond to the E-step
for truncated variational distributions.
One iteration of Eqns.\,\ref{EqnTVEMOptSteps} will be referred to as TV-EM iteration (as introduced in Sec.\,\ref{SecSummary}), and by definition the free energy is monotonically increased.

%, i.e., derivations of parameter update equations for a given generative model will be the same as for exact EM or other variational EM approaches. Only the expectation values
%these M-steps depend on 

\myvanish{After setting $\KKold=\KKnew$ and $\ThetaOld=\ThetaNew$
%After setting $\KK=\KKnew$, $\ThetaHat=\ThetaHatNew$, $\Theta=\ThetaNew$,
we start with the first optimization step again and obtain an
iterative procedure. We refer to one iteration step as a truncated
variational EM iteration (TV-EM iteration). The first two optimization
steps maximize the variational parameters $\ThetaHat$ and $\KK$, which
corresponds to the truncated variational E-step. The 
third optimization updates the model parameters is the M-step.
%
%While subspace-wise
%optimization is not necessarily as optimal as simulataneous
%optimization \comment{(citation here)}, it is the basis for any
%variational approach, and has proven successfull in a large number of
%applications.
%
By definition, a TV-EM iteration never decreases the free energy
$\FFHat(\KK,\ThetaHat,\Theta)$. Also partial optimization, i.e., improving
$\FFHat(\KK,\ThetaHat,\Theta)$ instead of maximizing it, in any optimization step does never decrease
the free energy. %For partial updates we, in general, typically also set $\theta$
}

The optimization steps (\ref{EqnTVEMOptSteps}) of TV-EM are formal definitions. In order
to be applicable in practice, a more concrete procedure for each of the three
optimization steps is required.\vspace{2mm}

%In the following we will turn to each optimization individually.

% ----------------------- THEORY PART -------------------------

\subsection{The Truncated Free Energy}
Instead of investigating and applying the three optimization steps
individually, we will carefully analyze each optimization in a
theoretically grounded way. For this purpose, let us first introduce a
free energy defined by setting the values of the variational
parameters $\ThetaHat$ equal to the model parameters $\Theta$:
%
%\small
\begin{equation}
\label{EqnFreeEnergyTVEM}
\hspace{-0mm}\ \FFHatHat(\KK,\Theta)\,:=\,\FFHat(\KK,\Theta,\Theta)\,=\, 
    \sum_{n=1}^{N} \bigg[
      \sum_{\sVec}\ \qn(\sVec;\KK,\Theta)\  
       \log\left( p(\sVec,\yVecN\,|\,\Theta)\right) 
      \bigg] 
      + H(\qn(\sVec;\KK,\Theta))\,.
\end{equation}
\normalsize
Given the definition of the truncated variational distribution $\qn(\sVec;\KK,\Theta)$ in (\ref{EqnQMain}), it can then be shown that
the free energy $\FFHatHat(\KK,\Theta)$ can be decisively simplified as follows:\\
\ \\[-1mm]
{\bf Proposition 3}\\*
Given a generative model defined by the joint distribution $p(\sVec,\yVec\,|\,\Theta)$. %, let $\LL(\Theta)$ be the likelihood in Eqn.\,\ref{EqnLikelihood}.
If $\FFHatHat(\KK,\Theta)$ is the free energy defined by (\ref{EqnFreeEnergyTVEM}) with truncated distributions given by (\ref{EqnQMain}), then it follows that\vspace{-2mm}
\begin{eqnarray}
\FFHatHat(\KK,\Theta) &=&  \sum_{n=1}^{N}\ \log\big(\sum_{\sVec\in\KKn} p(\sVec,\yVecN\,|\,\Theta)\ \big). \vspace{-2mm}
\label{EqnTruncatedF}
%
%\LL(\Theta)-\FFHatHat(\KK,\Theta) &=& D_{\mathrm{KL}}(...)\ =\ MAYBE NOT REQUIRED
%
\end{eqnarray}
\vspace{-5mm}\ \\
\noindent{\bf Proof}\\
Following Propositions 1 and 2, $\FFHatHat(\KK,\Theta)$ is a lower bound of $\LL(\Theta)$, which satisfies:\nopagebreak
\begin{eqnarray}
\disT\LL(\Theta)-\FFHatHat(\KK,\Theta)=\sum_{n}\DKL{\qn(\sVec;\KK,\Theta)}{p(\sVec\,|\,\yVecN,\Theta)}
\end{eqnarray}
For notational purposes let us introduce the normalizer $\Zn=\sum_{\sVec\in\KKn}p(\sVec\,|\,\yVecN,\Theta)$ such that:\vspace{-2mm}
\begin{eqnarray}
\qn(\sVec;\KK,\Theta) &=& \frac{1}{\Zn}\,p(\sVec\,|\,\yVecN,\Theta)\,\delta(\sVec\in\KKn)\vspace{-1mm}
\end{eqnarray}
From the above it follows that:\vspace{-1mm}
\small
\begin{eqnarray}
\lefteqn{\FFHatHat(\KK,\Theta)\ \ \ \ }\nonumber\\
 &=& \LL(\Theta)\,-\,\sum_{n}\DKL{\qn(\sVec;\KK,\Theta)}{p(\sVec\,|\,\yVecN,\Theta)}\nonumber\\[-1mm]
                  &=& \sum_{n}\log(p(\yVecN\,|\,\Theta))\,+\,\sum_{n}\sum_{\sVec} \qn(\sVec;\KK,\Theta) \log\big(\frac{p(\sVec\,|\,\yVecN,\Theta)}{\qn(\sVec;\KK,\Theta)}\big)\nonumber\\
                   &=& \sum_{n}\log(p(\yVecN\,|\,\Theta))\,+\,\sum_{n}\sum_{\sVec} \frac{1}{\Zn}p(\sVec\,|\,\yVecN,\Theta)\,\delta(\sVec\in\KKn) \log\big(\frac{p(\sVec\,|\,\yVecN,\Theta)}{\frac{1}{\Zn}p(\sVec\,|\,\yVecN,\Theta)\,\delta(\sVec\in\KKn))}\big)\nonumber\\
                   &=& \sum_{n}\log(p(\yVecN\,|\,\Theta))\,+\,\sum_{n}\sum_{\sVec\in\KKn} \frac{1}{\Zn}\,p(\sVec\,|\,\yVecN,\Theta) \log\big(\frac{p(\sVec\,|\,\yVecN,\Theta)}{\frac{1}{\Zn}p(\sVec\,|\,\yVecN,\Theta)}\big)\nonumber\\
                   &=& \sum_{n}\log(p(\yVecN\,|\,\Theta))\,+\,\sum_{n}\sum_{\sVec\in\KKn} \frac{1}{\Zn}\,p(\sVec\,|\,\yVecN,\Theta) \log\big(\Zn\big).
\end{eqnarray}
\normalsize
Again we used the convention $\qn(\sVec)\,\log\big( \qn(\sVec) \big) = 0$ for all $\qn(\sVec)=0$.
\begin{samepage}
Observing the intermediate result above, note that $\Zn$ is independent of $\sVec$. We can therefore continue as follows:\vspace{-1mm}
\small
\begin{eqnarray}
        \FFHatHat(\KK,\Theta) &=& \disS\sum_{n}\log(p(\yVecN\,|\,\Theta))\,+\,\sum_{n}\log\big(\Zn\big)  \frac{\sum_{\sVec\in\KKn}p(\sVec\,|\,\yVecN,\Theta)}{\Zn}\nonumber\\
                           &=& \sum_{n}\log(p(\yVecN\,|\,\Theta))\,+\,\sum_{n}\log\big(\Zn\big)\nonumber\\
                           &=& \sum_{n}\log(p(\yVecN\,|\,\Theta))\,+\,\sum_{n}\log\big(\sum_{\sVec\in\KKn}p(\sVec\,|\,\yVecN,\Theta)\big)\nonumber\\
                           &=& \sum_{n}\log(p(\yVecN\,|\,\Theta))\,+\,\sum_{n}\log\big(\frac{\sum_{\sVec\in\KKn}p(\sVec,\yVecN\,|\,\Theta)}{p(\yVecN\,|\,\Theta)}\big)\nonumber\\
                           &=& \sum_{n}\log\big(\sum_{\sVec\in\KKn} p(\sVec,\yVecN\,|\,\Theta)\ \big)\nonumber
%
%                   &=& \sum_{n}\log(p(\yVecN\,|\,\Theta))\,-\,\sum_{n}\DKL{\qn(\sVec;\KK,\Theta)}{p(\sVec\,|\,\yVecN,\Theta)}
%
\end{eqnarray}\normalsize
\vspace{-8mm}\ \\
%which proves the claim.\nopagebreak\\
\BOX\\
%\ \\
\end{samepage}
Considering Eqn.\,\ref{EqnTruncatedF} we instantly observe that $\FFHatHat(\KK,\Theta)$ is computationally tractable if $\KK$ is sufficiently small. Also the fact that $\FF(\KK,\Theta)$ lower-bounds the
log-likelihood can instantly be observed. Importantly, however, Prop.\,3 shows that Eqn.\,\ref{EqnTruncatedF} is a variational free energy which corresponds to the truncated variational distributions (\ref{EqnQMain}). Furthermore, Prop.\,3 directly relates (\ref{EqnTruncatedF}) to the free energy (\ref{EqnFreeEnergyTVEMOrg}). % and hence to the variational distributions (\ref{EqnQMain}) have to be proven, however.
As $\FFHatHat(\KK,\Theta)$ is a special case of $\FFHat(\KK,\ThetaHat,\Theta)$ both free energies are lower bounds of the likelihood $\LL(\Theta)$. Furthermore, it can be shown that $\FF(\KK,\ThetaHat,\Theta)$ can be obtained as a variational lower bound of $\FFHatHat(\KK,\Theta)$ and that the following holds.\\
\ \\
{\bf Proposition 4}\\
Given a generative model defined by the joint $p(\sVec,\yVec\,|\,\Theta)$, let $\LL(\Theta)$ be the likelihood in Eqn.\,\ref{EqnLikelihood}
and let $\FFHat(\KK,\ThetaHat,\Theta)$ and $\FFHatHat(\KK,\Theta)$ be the free energies defined by Eqn.\,\ref{EqnFreeEnergyTVEMOrg} and \ref{EqnTruncatedF}, respectively.
Then, for all values of $\KK$, $\ThetaHat$, and $\Theta$ the following applies:
\begin{eqnarray}
\LL(\Theta)\ \geq\ \FFHatHat(\KK,\Theta)\ \geq\ \FFHat(\KK,\ThetaHat,\Theta)\,.
\end{eqnarray}
{\bf Proof}\\
$\FFHatHat(\KK,\Theta)$ is a special case of $\FFHat(\KK,\ThetaHat,\Theta)$ by definition (Eqn.\,\ref{EqnFreeEnergyTVEM}), and as such a lower bound of $\LL(\Theta)$.
To show that $\FFHatHat(\KK,\Theta)\ \geq\ \FFHat(\KK,\ThetaHat,\Theta)$, we use Proposition 3 and apply Jensen's inequality:
\begin{eqnarray}
\FFHatHat(\KK,\Theta) &=& \sum_{n}\ \log\left(\sum_{\sVec\in\KKn} p(\sVec,\yVecN\,|\,\Theta)\right)\\
                 &=& \sum_{n}\ \log \left(\sum_{\sVec\in\KKn} \qhatn(\sVec) \frac{p(\sVec,\yVecN\,|\,\Theta)}{\qhatn(\sVec)} \right)\\
                  &\geq{}& \sum_{n}\sum_{\sVec\in\KKn}\ \qhatn(\sVec) \ \log \left( \frac{p(\sVec,\yVecN\,|\,\Theta)}{\qhatn(\sVec)} \right)\label{EqnFProof}
\end{eqnarray}
if $\sum_{\sVec\in\KKn}\qhatn(\sVec)=1$ and $\qhatn(\sVec)\geq{}0$ for all $\sVec\in\KKn$. We now define:
\begin{eqnarray}
 \qhatn(\sVec) &=& \frac{p(\sVec\,|\,\yVecN,\ThetaHat)}{\sum_{\sVec'\in\KKn}p(\sVec'\,|\,\yVecN,\ThetaHat)}\,.
\end{eqnarray}
The choice of $\qhatn(\sVec)$ fulfills the conditions for Jensen's inequality but note that it is not a probability density on the whole state space $\Omega$ of $\sVec$. Inserting $\qhatn(\sVec)$ into (\ref{EqnFProof}) we obtain:
\begin{eqnarray}
 \FFHatHat(\KK,\Theta) &\geq& \sum_{n}\sum_{\sVec\in\KKn}\ \frac{p(\sVec\,|\,\yVecN,\ThetaHat)}{\sum_{\sVec'\in\KKn}p(\sVec'\,|\,\yVecN,\ThetaHat)}
  \log\left( \frac{p(\sVec,\yVecN\,|\,\Theta)}{\frac{p(\sVec\,|\,\yVecN,\ThetaHat)}{\sum_{\sVec'\in\KKn}p(\sVec'\,|\,\yVecN,\ThetaHat)}}\right)\\
                    &=& \sum_{n}\sum_{\sVec}\  \qn(\sVec;\KK,\ThetaHat)\ \log\left( \frac{p(\sVec,\yVecN\,|\,\Theta)}{\qn(\sVec;\KK,\ThetaHat)}\right)\\
                    &=& \sum_{n}\sum_{\sVec}\  \qn(\sVec;\KK,\ThetaHat)\ \log\left(p(\sVec,\yVecN\,|\,\Theta)\right)\,+\,H(\qn(\sVec;\KK,\ThetaHat))\\
                    &=& \FFHat(\KK,\ThetaHat,\Theta)\,,
\end{eqnarray}
where $\qn(\sVec;\KK,\ThetaHat)$ is the truncated variational distribution in (\ref{EqnQMain}) and where $\FF(\KK,\ThetaHat,\Theta)$ is the corresponding free energy in (\ref{EqnFreeEnergyTVEMOrg}).\nopagebreak\\
\BOX\\
\ \\
%
%Based on Proposition 4 we now know that $\LL(\Theta)$, $\FFHatHat(\KK,\Theta)$, $\FFHat(\KK,\ThetaHat,\Theta)$ are a strictly ordered.
%
Having established that $\FFHat(\KK,\ThetaHat,\Theta)$ is a lower bound of $\FFHatHat(\KK,\Theta)$ for all $\ThetaHat$, the following applies for the differences
between $\LL(\Theta)$, $\FFHatHat(\KK,\Theta)$, and $\FFHat(\KK,\ThetaHat,\Theta)$:\\
\newcommand{\mms}{\hspace{-2mm}}
\ \\
{\bf Corollary 1}\\
All as above. 
\small
\begin{eqnarray}
 \LL(\Theta)-\FFHatHat(\KK,\Theta) \mms{}&=&\mms{} \sum_n\DKL{\qn(\sVec;\KK,\Theta)}{p(\sVec\,|\,\yVecN,\Theta)}\ \geq\ 0\,,\nonumber\\
 \FFHatHat(\KK,\Theta)-\FFHat(\KK,\ThetaHat,\Theta)\mms{}&=&\mms{} \sum_n\DKL{\qn(\sVec;\KK,\ThetaHat)}{p(\sVec\,|\,\yVecN,\Theta)}\,-\,\sum_n\DKL{\qn(\sVec;\KK,\Theta)}{p(\sVec\,|\,\yVecN,\Theta)}\geq\ 0\,,\nonumber\\
 \LL(\Theta)-\FFHat(\KK,\ThetaHat,\Theta) \mms{}&=&\mms{} \sum_n\DKL{\qn(\sVec;\KK,\ThetaHat)}{p(\sVec\,|\,\yVecN,\Theta)}\geq\ 0\,.\vspace{-3mm}\nonumber
\end{eqnarray}
\normalsize
{\bf Proof}\\*
The first and the last equation are a direct consequence of Propositions~2 and 4 and of the fact that 
$\FFHatHat(\KK,\Theta)$ and $\FFHat(\KK,\ThetaHat,\Theta)$ are both variational
lower bounds of $\LL(\Theta)$. The second equation is obtained by taking the difference of the
first and the last equation.  The fact that the difference of the
KL-divergences of the second equation is greater zero follows from
Proposition 4.\nopagebreak\\
\BOX\\[3mm]
%\ \\
By considering Corollary~1, we can now solve the last optimization step (Opt~3)
in (\ref{EqnTVEMOptSteps}) analytically. Indeed, it can be shown that $\FFHat(\KK,\ThetaHat,\Theta)$
is optimized w.r.t.\ $\ThetaHat$ if the values of the variational parameters $\ThetaHat$ are set equal to model parameters $\Theta$:\\[6mm]
%\ \\
{\bf Proposition 5}\\
If $\FFHat(\KK,\ThetaHat,\Theta)$ is the truncated free energy of Eqn.\,\ref{EqnFreeEnergyTVEMOrg} and if $\KK$ and $\Theta$ are fixed,
then $\ThetaHat=\Theta$ is a global maximum of $\FFHat(\KK,\ThetaHat,\Theta)$. 
%
%it applies for fixed $\KK$ and $\Theta$ that
%$\ThetaHat=\Theta$ is a global maximum of $\FFHat(\KK,\ThetaHat,\Theta)$.
%

\ \\[2mm]
\noindent{\bf Proof}\\*
Let us first re-express $\FFHat(\KK,\ThetaHat,\Theta)$ using Corollary~1:
\footnotesize
\begin{eqnarray}
\FFHat(\KK,\ThetaHat,\Theta) &=& \FFHatHat(\KK,\Theta)\,+\,\sum_n\DKL{\qn(\sVec;\KK,\Theta)}{p(\sVec\,|\,\yVecN,\Theta)}\,-\,\sum_n\DKL{\qn(\sVec;\KK,\ThetaHat)}{p(\sVec\,|\,\yVecN,\Theta)}\nonumber\\
                             &=& \FFHatHat(\KK,\Theta)\,+\,\sum_n\DKL{\qn(\Theta)}{\pn(\Theta)}\,-\,\sum_n\DKL{\qn(\ThetaHat)}{\pn(\Theta)},\normalsize
\label{EqnFreeEnergyProofTemp}
\end{eqnarray}
\normalsize
where the last line abbreviates the distributions for readability.
As only the last summand depends on $\ThetaHat$, $\FFHat(\KK,\ThetaHat,\Theta)$ is maximized if
$\sum_n\DKL{\qn(\ThetaHat)}{\pn(\Theta)}$ is minimized. 
Now we also know from Corollary~1 that $\sum_n\DKL{\qn(\ThetaHat)}{\pn(\Theta)}\geq{}\sum_n\DKL{\qn(\Theta)}{\pn(\Theta)}$,
where only the left-hand-side depends on $\ThetaHat$. Hence, the most minimal value for $\sum_n\DKL{\qn(\ThetaHat)}{\pn(\Theta)}$ achievable is 
%\footnotesize
\begin{eqnarray}
\phantom{iiiiiiiii}\disT\sum_n\DKL{\qn(\ThetaHat)}{\pn(\Theta)}\,=\,\sum_n\DKL{\qn(\Theta)}{\pn(\Theta)}\,.
\end{eqnarray}
\normalsize
By choosing $\ThetaHat=\Theta$ we can indeed satisfy the equality, and therefore know that\linebreak \mbox{$\sum_n\DKL{\qn(\ThetaHat)}{\pn(\Theta)}$}
takes on a global minimum for this choice. This global minimum then implies (because of Eqn.\,\ref{EqnFreeEnergyProofTemp}) a global maximum of $\FFHat(\KK,\ThetaHat,\Theta)$ w.r.t.\,$\ThetaHat$.
\nopagebreak\\
\BOX\\
\ \\
Note that there can potentially be other global maxima, e.g., due to
permutations of the parameters without effect on
$\qn(\sVec;\KK,\ThetaHat)$. Using the KL-divergences of Corollary~1
makes it salient that it is sufficient to equate
$\qn(\sVec;\KK,\ThetaHat)$ and $\qn(\sVec;\KK,\Theta)$,
which is a weaker condition than equating $\ThetaHat$ and $\Theta$.
To show that $\ThetaHat=\Theta$ is maximizing
$\FF(\KK,\ThetaHat,\Theta)$ we can, alternatively, also use Proposition 4 directly, i.e., $\FFHatHat(\KK,\Theta)\geq{}\FFHat(\KK,\ThetaHat,\Theta)$.
Either way, we can conclude that $\FFHat(\KK,\ThetaHat,\Theta)$ is maximized for $\ThetaHat=\Theta$, in which case $\FFHat(\KK,\ThetaHat,\Theta)$ becomes equal to $\FFHatHat(\KK,\Theta)$.

Using\vspace{-2mm} the customary `$\argmax{}$' notation for variational optimization, Prop.\,5 can more informally be stated as
\begin{eqnarray}
%
%	\Theta &=& \argmax{\ThetaHat}\big\{\FFHat(\KK,\ThetaHat,\Theta)\big\}\,.
	\argmax{\ThetaHat}\big\{ \FFHat(\KK,\ThetaHat,\Theta) \big\} &=& \Theta\,.\label{EqnThirdOpt}
\end{eqnarray}
%
%while keeping in mind potentially multiple maxima. Considering \refp{EqnThirdOpt}, 
%
Prop.\,5 thus allows for solving the third optimization step of Eqns.\,\ref{EqnTVEMOptSteps}, and consequently one
TV-EM iteration reduces to two optimizations: by applying (\ref{EqnThirdOpt}) the third optimization (Opt~3 of Eqns.\,\ref{EqnTVEMOptSteps}), is simply given
by $\ThetaHatNew=\ThetaNew$. After combining this update with the last line of Eqns.\,\ref{EqnTVEMOptSteps} we obtain:
%
%first optimization step provided by Proposition~5,
%the iterative TV-EM of procedure of Eqn.\,\ref{EqnTVEMOptSteps} now reduces to essentially
%two optimization steps:
%
%
\begin{equation}
\label{EqnOptFinalPre}
\begin{array}{lcllll}
\mathrm{Opt\ 1:} & &    \KKnew &= \hspace{1mm}\argmax{\KK}\big\{\FFHat(\KK,\ThetaHatOld,\ThetaOld)\big\} & & \mbox{}\hspace{-6mm}\phantom{\int^f_g}\\
\mathrm{Opt\ 2:} & &  \ThetaNew  &=\hspace{1mm}\argmax{\Theta}\big\{\FFHat(\KKnew,\ThetaHatOld,\Theta)\big\} & & \mbox{}\hspace{-6mm}\phantom{\int^f_g}\\[3mm]
%
%\mathrm{} & &         \ThetaHatNew &= \hspace{1mm}\argmax{\ThetaHat}\big\{\FFHat(\KKnew,\ThetaHat,\ThetaNew)\big\} & & \mbox{while holding $\KKnew$ and $\ThetaNew$ fixed}\hspace{-6mm}\phantom{\int^f_g}\\[2mm]
%
\mathrm{} & &          \ThetaHatOld &\hspace{0mm}=\hspace{2mm} \ThetaNew \mbox{\ \ \ and\ \ \ } \ThetaOld\hspace{2mm}=\hspace{2mm}\ThetaNew
\end{array}
\end{equation}
As the variational parameters $\ThetaHatOld$ and the model parameters $\ThetaOld$ are now both set to the same values in the last line of (\ref{EqnOptFinalPre}), we can now for Opt\,1 use (without loss of generality) the simplified form of free energy $\FF(\KK,\Theta)$ as given by Eqn.\,\ref{EqnTruncatedF}. As a consequence, we can replace the two resets of the parameters $\Theta$ in the last line by
the single reset  $\ThetaOld=\ThetaNew$ and finally obtain the TV-EM formulation of Eqns.\,\ref{EqnTVEStep} to \ref{EqnTVThird} as stated in the beginning. % (Eqns.\,\ref{EqnTVEStep} to \ref{EqnTVThird}).

To recapitulate, we have thus finally proven that iterating the TV-EM steps of Eqns.\,\ref{EqnTVEStep} to \ref{EqnTVThird} monotonically increases the free energy (\ref{EqnFreeEnergyTVEMOrgA}), which
is a lower bound of the log-likelihood (\ref{EqnLikelihood}). The proof follows from the updates (\ref{EqnTVEMOptSteps}), which monotonically increase the free energy by definition. Using Propositions~1 to 5 and Corollary 1, Eqns.\,\ref{EqnTVEStep} to \ref{EqnTVThird} result in the same updates as (\ref{EqnTVEMOptSteps}) but represent a strong simplification. Note, in this respect, that it is important for the TV-M-step (\ref{EqnTVMStep}) to be of the same form as for exact EM and other
types of variational EM approximations because this form means that any M-step equations derived for any previously considered generative model can be reused for truncated variational EM. 
Maintaining the standard M-step update is non-trivial for truncated distributions and required the application of the theoretical results derived above: 

First, we needed to start with a three-stage optimization for $\FF(\KK,\ThetaHat,\Theta)$. A direct optimization of the simplified free energy $\FF(\KK,\Theta)$ in a two-stage procedure would change the M-step to a non-standard form. This is because the truncated distributions do also depend on $\Theta$ (and derivatives w.r.t.\ to all $\Theta$'s would be required). By treating
$\ThetaHat$ as variational parameters, derivatives exclusively w.r.t.\ the log-joint probability of the generative model can be taken, and this results, e.g., in the well-known closed-form updates of Gaussian Mixture Models, Hidden Markov Models, Factor Analysis etc. All these M-step results and similar such results for many other models can thus directly be used for TV-EM.

Second, when we take derivatives w.r.t.\ $\Theta$ in the TV-M-step, it may feel straight-forward to hold the parameters $\Theta$ of the variational distributions fixed at
their old values $\ThetaOld$; and to only afterwards
set $\ThetaOld=\ThetaNew$ as in (\ref{EqnTVThird}). We are very used to this procedure for exact EM. Note, however, that for exact EM, it is required to prove that such a procedure never decreases the free energy. A possible such proof for exact EM would use Eqn.\,\ref{EqnDKLStandard} and full posteriors $p(\sVec\,|\,\yVecN,\ThetaHat)$ as variational distributions (using $\ThetaHat$ as variational parameters). The KL-divergence
$\DKL{p(\sVec\,|\,\yVecN,\ThetaHat)}{p(\sVec\,|\,\yVecN,\Theta)}$ can then be set to zero by choosing $\ThetaHat=\Theta$, which according to Eqn.\,\ref{EqnDKLStandard} globally maximize the free energy
\citep[also see Lemma~1 of][]{NealHinton1998}. For general variational distributions with $\ThetaHat$ as variational parameters, the same does {\em not} apply. For truncated variational distributions, it is Prop.\,5 which shows that we can proceed with truncated distributions in the same way as we are used to for full posteriors in exact EM. Indeed, Prop.\,5 contains Lemma~1 of \citet[][]{NealHinton1998} as a special case: If we set all $\KKn$ to contain all states $\sVec$, i.e.\, $\KKn=\Omega$ for all $n$, then the variational distributions $\qn(\sVec;\KK,\Theta)$ in (\ref{EqnQMain}) become equal to the full posteriors. Consequently, the free energy $\FF(\KK,\ThetaHat,\Theta)$ becomes equal to the standard free energy for exact EM, and Prop.\,5 then shows that this free energy is maximized if we set $\ThetaOld$ of the posteriors $p(\sVec\,|\,\yVecN,\ThetaOld)$ equal to the $\Theta$ obtained in the M-step. Also note that for $\KKn=\Omega$, the TV-EM algorithm reduces to exact EM as the TV-E-step (\ref{EqnTVEStep}) becomes trivial
and as the truncated distributions (\ref{EqnQMain}) become equal to the exact posteriors.
%
%Finally, another way to regard, e.g., Props.\,3 is to notice that the simplified free energy does not contain an entropy term. All non-trivial variational distributions
%(including mean-field or Gaussians) have a finite entropy, and the entropy is important to determine the variational parameters. Only for standard EM (no variational parameters) or for `hard EM' approaches
%(zero entropy), an entropy term is trivially not important. Props.\,3 now shows that the free energy of the non-trivial truncated distributions can be rewritten without
%and entropy term. 

\subsection{Partial Truncated E- and M-Steps}
\label{SecPartialTVEM}
So far, we have considered with Eqns.\,\ref{EqnTVEMOptSteps} full maximizations of truncated free energies, for which we derived the TV-EM algorithm given by Eqns.\,\ref{EqnTVEStep} to \ref{EqnTVThird}.
However, for many generative models such full maximizations are analytically and/or computationally intractable. In order to also address these important cases, we here apply
Props.\,~1 to 5 to partial TV-E- and partial TV-M-steps (which is analogous but not equal to the full maximization). Let us start with a three-stage optimization as before but instead we now consider a partial TV-EM procedure:
\begin{equation}
\begin{array}{lcllll}
\label{EqnTVEMOptStepsParial}
\mathrm{Opt\ 1:} & &   \mbox{choose $\KKnew$ such that\phantom{iiiiii}} \FFHat(\KKnew,\ThetaHatOld,\ThetaOld) &\geq\hspace{2mm} \FFHat(\KKold,\ \ThetaHatOld,\ThetaOld)\hspace{-6mm}\phantom{\int^f_g}\\
\mathrm{Opt\ 2:} & &   \mbox{choose $\ThetaNew$ such that\phantom{iiiiii}} \FFHat(\KKnew,\ThetaHatOld,\ThetaNew) &\geq\hspace{2mm} \FFHat(\KKnew,\ThetaHatOld,\ThetaOld)\hspace{-6mm}\phantom{\int^f_g}\\
\mathrm{Opt\ 3:} & &   \ThetaHatNew = \hspace{1mm}\argmax{\ThetaHat}\big\{\FFHat(\KKnew,\ThetaHat,\ThetaNew)\big\} \hspace{-6mm}\phantom{\int^f_g}\\[2mm]
\mathrm{} & &          \KKold \hspace{0mm}=\hspace{2mm} \KKnew, \mbox{\ \ } \ThetaOld\hspace{2mm}=\hspace{2mm}\ThetaNew, \mbox{\ \ } \ThetaHatOld \hspace{0mm}=\hspace{2mm} \ThetaHatNew
\end{array}
\end{equation}
An iteration using (\ref{EqnTVEMOptStepsParial}) monotonically increases the free energy $\FFHat(\KK,\ThetaHat,\Theta)$ as each individual optimization by definition never decreases $\FFHat(\KK,\ThetaHat,\Theta)$.
Opt~1 and Opt~2 are now partial optimizations ($\FFHat(\KK,\ThetaHat,\Theta)$ is increased, not maximized), while we maintain for Opt~3 a full maximization. By applying Prop.\,5 we can now, as before, replace
the maximization of Opt~3 by setting $\ThetaHatNew=\ThetaNew$. Also as before, we then obtain by combining the analytical solution of Opt~3 with the last line of (\ref{EqnTVEMOptStepsParial}) a two-stage optimization procedure:
\begin{equation}
\begin{array}{lcllll}
\label{EqnTVEMOptStepsParialTemp}
\mathrm{Opt\ 1:} & &   \mbox{choose $\KKnew$ such that\phantom{iiiiii}} \FFHat(\KKnew,\ThetaHatOld,\ThetaOld) &\geq\hspace{2mm} \FFHat(\KKold,\ \ThetaHatOld,\ThetaOld)\hspace{-6mm}\phantom{\int^f_g}\\
\mathrm{Opt\ 2:} & &   \mbox{choose $\ThetaNew$ such that\phantom{iiiiii}} \FFHat(\KKnew,\ThetaHatOld,\ThetaNew) &\geq\hspace{2mm} \FFHat(\KKnew,\ThetaHatOld,\ThetaOld)\hspace{-6mm}\phantom{\int^f_g}\\
\mathrm{} & &          \KKold \hspace{0mm}=\hspace{2mm} \KKnew, \mbox{\ \ } \ThetaOld\hspace{2mm}=\hspace{2mm}\ThetaNew, \mbox{\ \ } \ThetaHatOld \hspace{0mm}=\hspace{2mm} \ThetaNew
\end{array}
\end{equation}
As $\ThetaOld$ and $\ThetaHatOld$ are now set to the same values, we can replace the free energy $\FFHat(\KK,\ThetaHat,\Theta)$ of Opt~1 in (\ref{EqnTVEMOptStepsParialTemp}) by the simplified free energy (\ref{EqnTruncatedF}) derived for Proposition~3. By further simplifying the last line of (\ref{EqnTVEMOptStepsParialTemp}) we finally arrive at:
\begin{equation}
\begin{array}{lcllll}
\label{EqnTVEMOptStepsParialFinal}
\mathrm{Opt\ 1:} & &   \mbox{choose $\KKnew$ such that\phantom{iiiiii}} \phantom{iiiiiik}\FFHat(\KKnew,\ThetaOld) &\geq\hspace{2mm} \FFHat(\KKold,\ \ThetaOld)\hspace{-6mm}\phantom{\int^f_g}\\
\mathrm{Opt\ 2:} & &   \mbox{choose $\ThetaNew$ such that\phantom{iiiiii}} \FFHat(\KKnew,\ThetaOld,\ThetaNew) &\geq\hspace{2mm} \FFHat(\KKnew,\ThetaOld,\ThetaOld)\hspace{-6mm}\phantom{\int^f_g}\\
\mathrm{} & &          \KKold \hspace{0mm}=\hspace{2mm} \KKnew, \mbox{\ \ } \ThetaOld\hspace{2mm}=\hspace{2mm}\ThetaNew
\end{array}
\end{equation}
Eqns.\,\ref{EqnTVEMOptStepsParialFinal} will be referred to as a partial TV-EM iteration. Like non-partial TV-EM (i.e., Eqns.\,\ref{EqnTVEStep} to \ref{EqnTVThird}), a partial TV-EM step monotonically increases
the truncated free energy (\ref{EqnTruncatedF}). While the derivation of partial TV-EM used the same theoretical results as non-partial TV-EM, note that a main difference is that the variational parameters $\KK$
have to be memorized across partial TV-EM iterations. A full optimization does not necessarily require such a memorization. Furthermore, we require initial values of $\KK$ for partial TV-EM.
Finally, observe that we can, in the same way as above, define TV-EM algorithms with only the TV-E-step being a partial optimization or with only the TV-M-step being a partial optimization.

\subsection{Explicit Form}
\label{SecExplicitForm}
Before we consider applications of the theoretical results for TV-EM, let us formulate the algorithm given in the previous section more explicitly. 
Obtaining an explicit form for the TV-E-step is straight-forward by just inserting the simplified truncated free energy (\ref{EqnTruncatedF}) into
the first optimization of Eqns.\,\ref{EqnTVEMOptStepsParialFinal} (see further below).
Regarding the TV-M-step, consider the second optimization of Eqns.\,\ref{EqnTVEMOptStepsParialFinal} and let us insert the
(non-simplified) truncated free energy (\ref{EqnFreeEnergyTVEMOrg}). After noting that (as usual for variational approaches) the entropy term
is not relevant for the optimization of model parameters $\Theta$, the relevant function to optimize is given by:
\begin{eqnarray}
\label{EqnQFunction}
% %
      \backOne\backOne Q(\Theta)\backHalf &=&\backHalf \sum_{n=1}^{N} \sum_{\sVec}\ \qn(\sVec;\KK,\ThetaOld)\ \log\left( p(\sVec,\yVecN\,|\,\Theta)\right) = \sum_{n=1}^{N}\E{\log\big( p(\sVec,\yVecN\,|\,\Theta)\big)}_{\qn(\sVec;\KK,\ThetaOld)}
% %
\end{eqnarray}
If we now insert Eqn.\,\ref{EqnSuffStatMain} for the expectation value w.r.t.\,$\qn(\sVec;\KK,\ThetaOld)$ in (\ref{EqnQFunction}), we obtain (together with the TV-E-step) an explicit form of one TV-EM iteration given by:
\begin{equation}
\begin{array}{llcll}
\label{EqnTVEMExplicit}
\hspace{-3mm}\mbox{\bf TV-E-step:} \\
    \mbox{\ \ change $\KK$ from $\KKoldsub$ to $\KKnewsub$ such that}   &  \disS\sum_{n}\sum_{\sVec\in\KKn}\ \log\left( p(\sVec,\yVecN\,|\,\ThetaOld)\right)\phantom{i}    \mbox{increases.} \hspace{-6mm}\phantom{\int^f_g}\\[9mm]
\hspace{-3mm}\mbox{\bf TV-M-step:}\\[-7mm]
    \mbox{\ \ change $\Theta$ from $\ThetaOld$ to $\ThetaNew$ such that}     &  
{\scriptstyle \disS\sum_{n}\ \frac{\disS\sum_{\sVec\in\KKnewN} p(\sVec,\yVecN\,|\,\ThetaOld)\log\left( p(\sVec,\yVecN\,|\,\Theta)\right) }{\disS\sum_{\sVecPrime\in\KKnewN}p(\sVecPrime,\yVecN\,|\,\ThetaOld)}}\phantom{i}              \mbox{increases.} \\[8mm]
\ \\
\hspace{-3mm}\mbox{\bf Reset:} &            \KKoldsub \hspace{0mm}=\hspace{2mm} \KKnewsub, \mbox{\ \ } \ThetaOld\hspace{2mm}=\hspace{2mm}\ThetaNew\hspace{-9mm}\vspace{0mm}\ \\[-14mm]
\end{array}
\end{equation}
\ \\[5mm]
%To obtain the explicit form of the TV-E-step, we simply inserted the simplified truncated free energy (\ref{EqnTruncatedF}) into the first optimization of Eqns.\,\ref{EqnTVEMOptStepsParialFinal}.

The form (\ref{EqnTVEMExplicit}) of one TV-EM iteration makes the following explicit: (A)~The procedure is fully defined by the joint probability of the considered generative model; and (B)~all entities that have to be computed are computationally tractable given sufficiently small sets $\KKn$ and efficiently computable joint probabilities. Eqns.\,\ref{EqnTVEMExplicit} also highlight the very concise form of the procedure.

If TV-EM is applied to a given generative model, the M-step typically relies on derivatives of the log-joint probability (be it either to derive closed-form update equations or gradient equations for more intricate models). The E-step will ultimately reduce to a pair-wise comparison of joint probabilities, which can be realized efficiently (we will provide more details of such procedures in the following).
% The next section will provide more details.
%
% (see next section for details-- and we will discuss some examples. In the following we will discuss some concrete examples.
%
%\newpage
%
%\newpage
%
%\newpage
%
%\newpage
%
\section{Applications of Theoretical Results}
\label{SecApplications}
Our theoretical results and the TV-EM meta-algorithm can now be applied to provide novel theoretical insights,
and to point to novel ways to develop learning algorithms for generative models.
We will consider three applications of our results: First, we will consider multiple-cause or latent variable generative models, i.e.,
models for which the values of multiple latent variables combine to generate the data. Second, we consider applications to
mixture models, i.e., models for which an observation is always generated by one latent variable. Finally, in the third application,
we investigate the relation between TV-EM and `hard EM'. The latter being a very widely applied and in practice very successful form of
parameter optimization in generative models.
%
%We will consider three applications of our results with the goal of providing novel theoretical insights, and to point to novel ways to develop learning algorithms for generative models.
%First, we will consider multiple-cause generative models with large combinatorial hidden spaces, and show how TV-EM can be applied to derive computationally tractable algorithms for
%
%The variations point to novel algorithmic approaches 
%and connect to previous algorithms that were developed based on truncated approaches without using the novel theoretical results obtained in this study. 
%
%
%
\subsection{TV-EM for Multiple-Cause Models}
\label{SecSampling}
Multiple-cause models or latent variable models are generative models in which a set of latent variables (latent causes) combine to generate an observation.
Common examples are sparse coding models \citep[e.g.][]{OlshausenField1996}, noisy-OR Bayes nets \citep[e.g.][]{SingliarHauskrecht2006} or sigmoid
believe networks \citep[][]{SaulEtAl1996,JordanEtAl1999}. 
% but in general any directed graphical model (including deep versions) can be considered a multiple-cause model if multiple causes can combine to generate observed variables.
The interaction of multiple hidden variables typically gives rise to large latent spaces because of the combinatorics of individual latents.
For any larger models, full posteriors over the latent space are not computationally tractable anymore, and variational EM is a standard technique
to address such intractabilities.

%In contrast to common variational approaches such as mean-field, TV-EM uses sets of hidden states as variational parameters.
%This property provides a natural link between TV-EM and sampling procedures, which also use sets of latent states to obtain tractable approximations.
For our application to multiple-cause models let us consider partial TV-EM (Eqns.\,\ref{EqnTVEMOptStepsParialFinal}).
Similar to exact EM, the M-step of TV-EM requires the computation of derivatives of the log-joint probability of the considered generative model
(compare Eqns.\,\ref{EqnTVEMExplicit}). Computing any such derivations is standard except of the expectation values
w.r.t.\ $\qn(\sVec;\KK,\Theta)$ which are for TV-EM computed using (\ref{EqnSuffStatMain}). The crucial difference to previous variational approaches
is hence the truncated variational E-step. Instead of solving fixed-point equations for the variational parameters as, e.g., for mean-field
approaches, we have to find variational parameters that take the form of finite sets of hidden states. We will term these
states {\em variational states}. A (partial) TV-E-step can now be implemented by suggesting new variational states $\KKtilde$ and to compare
the free energy (\ref{EqnTruncatedF}) of these new states with the free energy of the old variational states. The set $\KK$ is then replaced
by a new set $\KKtilde$ if the free energy increases. %, i.e., if a partial TV-E-step is realized. 
The efficiency of this procedure will, of course,
depend crucially on the way how new variational states are suggested or how new sets $\KK$ are defined. Before we consider more concrete
examples of TV-E-steps, let us formulate the above described procedure (which directly follows form Eqns. \ref{EqnTVEMOptStepsParialFinal}) as a
meta-algorithm (see Alg.\,1).
%

%\newcommand{\algBreak}{\vspace{1.5mm}\\}
%
%\incmargin{0.0em}
%\restylealgo{boxed}
%\linesnumbered
%
%\paragraph{Blind sampling.}  
\newcommand{\algBreak}{\vspace{1.5mm}\\}
\begin{algorithm}[h]\vspace{1.5mm}
%\dontprintsemicolon
init model parameters $\ThetaOld$;\algBreak
init variational states $\KKold$;\algBreak
\Repeat{parameters $\Theta$ have sufficiently converged\vspace{1.5mm}}
{
%set $\ThetaHat=\Theta$\\%\algBreak
%
\Repeat{$\FFHatHat(\KKnew,\ThetaOld)$ has sufficiently increased\vspace{2mm}}{
suggest new set of states $\KKtilde$;\algBreak
%given $\KKold$ sample new variational states to define $\KKtilde$;\algBreak
%
% sample new variational states $\KKtilde$
%
%compute a variation $\KKtilde$ of $\KKold$;\algBreak
%
\If{$\FFHatHat(\KKtilde,\ThetaOld)>\FFHatHat(\KKold,\ThetaOld)$}{$\KKnew=\KKtilde$}
}
compute $\ThetaNew$ that optimizes $\disS Q(\Theta)=\sum_{n=1}^{N}\E{\log\big( p(\sVec,\yVecN\,|\,\Theta)\big)}_{\qn(\sVec;\KKnew,\ThetaOld)}$;\\
$\KKold = \KKnew$;\algBreak
$\ThetaOld = \ThetaNew$;\algBreak
}
%--------------------------------------------------------------------------------------------------------------------\\[2mm]
%
%The functions $\FFHatHat(\KK,\Theta)$ is given by
%
%small
%$\disS\FFHatHat(\KK,\Theta)\,=\,\sum_{n}\ \log\big(\sum_{\sVec\in\KKn}\ p(\sVec,\yVecN\,|\,\Theta)\big)\,,$\ see (\ref{EqnTruncatedF}).\\
%
\normalsize
% $
% \begin{array}{rcl}
% \qn(\sVec;\KK,\Theta) &=& \frac{\disT\phantom{\int}p(\sVec,\yVecN\,|\,\Theta)\phantom{\int}}
% {\disT\sum_{\sVecT\in\KKn}p(\sVecT,\yVecN\,|\,\Theta)}\,\delta(\sVec\in\KKn)\\[1mm]
% %
% \FFHatHat(\KK,\Theta) &=& \sum_{n}\ \log\big(\sum_{\sVec\in\KKn}\ p(\sVec,\yVecN\,|\,\Theta)\big)
% %
% \end{array}
% $
%
\caption{Application to Multiple-Cause Models.\label{TabET}}
\end{algorithm}
\myvanish{
\vspace{-81mm}
\begin{equation}
\phantom{mmmmmmmmmmmmmmmmmmmmmmmmmmmmmmmmmmmmmmmmmmmm}
\left.
\begin{array}{c}
 \\
 \\
 \\[2mm]
 \\

\end{array}
\right] \phantom{}\mbox{{\bf TV-E-step}}\nonumber
\end{equation}
\vspace{-1mm}
\begin{equation}
\phantom{mmmmmmmmmmmmmmmmmmmmmmmmmmmmmmmmmmmmmmmmmmiiii}
\left.
\begin{array}{c}
 \\[1mm]
 \\

\end{array}
\right] \phantom{}\mbox{{\bf TV-M-step}}\nonumber
\end{equation}
\vspace{39mm}
}
%

%One way to implement a partial TV-E-step is to vary the states in $\KK$ and to select those
%states that improve the simplified free energy $\FFHat(\KK,\ThetaOld)$ while $\ThetaOld$ is kept fixed. A straight-forward option
%to vary the states in $\KK$ is the use of sampling procedures, and the way how $\KK$ is varied using sampling will decide about the
%efficiency of TV-EM optimization procedure. 
%
%Before we consider concrete sampling procedures, let us formulate a meta-algorithm which directly follows from partial
%TV-EM (Eqns. \ref{EqnTVEMOptStepsParialFinal}) and which uses a sampler
%to vary the states in $\KK$.

%
The inner loop of Alg.\,1 repeatedly changes the states in $\KK$ and its if-clause warrants that $\FFHatHat(\KK,\Theta)$ is
increased w.r.t.\,$\KK$.
The successive optimization of $\FFHatHat(\KK,\Theta)$ w.r.t.\,$\Theta$ can be accomplished using standard M-step update
equations with expectation values estimated by (\ref{EqnSuffStatMain}). Together with setting $\KKold = \KKnew$ and $\ThetaOld = \ThetaNew$,
one iteration thus realizes one partial TV-EM step. The outer loop of Alg.\,1 iterates over individual TV-EM steps such that
Alg.\,1 provably monotonically increases the free energy $\FFHatHat(\KK,\Theta)$.
As it is the case for the parameters $\Theta$ of the outer loop, we can
terminate the inner loop (the TV-E-step) after one iteration (if the free energy increased) or
once there are no or no significant changes of $\KK$ observed anymore, or at any intermediate step.
%It may, however, make more sense to terminate the inner loop earlier once $\FFHatHat(\KK,\Theta)$
%has increased sufficiently. 
%The best strategy will presumably be model and data dependent.%type the TV-EM algorithm is applied to.

Considering Alg.\,1, the computation and comparison of free energies may seem computationally demanding. 
However, thanks to the specific functional form of the simplified free energy (\ref{EqnTruncatedF}), 
it is sufficient to pair-wise compare the joint probabilities instead having to compute the whole free energies. That is, 
for a given data point $\yVecN$ and any newly suggested state, it is sufficient to compare the
joint probability of the new state with those of the variational states in $\KKn$:

\ \\
{\bf Proposition 6}\\*
Consider the application of TV-EM to a generative model given by the joint $p(\sVec,\yVec\,|\,\Theta)$ % mixture model given by (\ref{EqnMixtureModel}), % and TV-EM , its truncated free energy of Eqn.\,\ref{} (with $\sVec$ replaced by $c$),
and let $\KKn$ be the set of variational states for a data point $\yVecN$.  
If we now replace a state $\sVecOld$ in $\KKn$ by a new state $\sVecNew$ so far not in $\KKn$ then 
the free energy $\FF(\KK,\Theta)$ of (\ref{EqnTruncatedF}) is increased if and only if 
\begin{equation}
\label{EqnCriterion}
p(\sVecNew,\yVecN\,|\,\Theta)\ >\ p(\sVecOld,\yVecN\,|\,\Theta).
\end{equation}
{\bf Proof}\\[1mm]
For a given data point $n$ let us consider $\KKn$ and let $\KKnewsub^{(n)}$ be the set defined by replacing the latent state $\sVecOld\in\KKn$ with $\sVecNew\not\in\KKn$, i.e., $\KKnewsub^{(n)}=\KKn \backslash \{\sVecOld\} \cup \{\sVecNew\}$.
Let us further define $\KKnew$ by replacing $\KKn$ with the set $\KKnewsub^{(n)}$, i.e., $\KKnew=\{\KK^{(1)},\ldots,\KK^{(n-1)},\KKnewsub^{(n)}, \KK^{(n+1)},\ldots,\KK^{(N)}$.
Then it follows:\vspace{5mm}

$
\begin{array}{lrcl}
& \FF(\KKnew,\Theta) &>& \FF(\KK,\Theta)\\[2mm]
\Leftrightarrow{} & \hspace{-0mm}\disS\sum_{\underset{m\neq{}n}{m=1}}^{N}\log\big(\backHalf\sum_{\ \sVec\in\KK^{(n)}} p(\sVec,\yVec^{(n)}\,|\,\Theta)\,\big)\,+\,\log\big(\backHalf\sum_{\ \sVec\in\KKnewsub^{(n)}} p(\sVec,\yVecN\,|\,\Theta)\,\big)\hspace{-45mm}\\[2mm]
              &&  \hspace{-50mm} >\ \ \disS\sum_{\underset{m\neq{}n}{m=1}}^{N}\log\big(\backHalf\sum_{\ \sVec\in\KK^{(n)}} p(\sVec,\yVec^{(n)}\,|\,\Theta)\,\big)\,+\,\log\big(\backHalf\sum_{\ \sVec\in\KKn} p(\sVec,\yVecN\,|\,\Theta)\,\big)\hspace{-40mm}\\[8mm]
\Leftrightarrow{}& \disS\log\big(\backHalf\sum_{\ \sVec\in\KKnewsub^{(n)}} p(\sVec,\yVecN\,|\,\Theta)\,\big)   &>& \disS    \log\big(\backHalf\sum_{\ \sVec\in\KKn} p(\sVec,\yVecN\,|\,\Theta)\,\big)\\[4mm]
\Leftrightarrow{}& \disS\sum_{\underset{\sVec\neq{}\sVecNew}{\sVec\in\KKn}} p(\sVec,\yVecN\,|\,\Theta)\,+\,p(\sVecNew,\yVecN\,|\,\Theta)   &>& \disS  \sum_{\underset{\sVec\neq{}\sVecNew}{\sVec\in\KKn}} p(\sVec,\yVecN\,|\,\Theta)\,+\,p(\sVecOld,\yVecN\,|\,\Theta)\\[4mm]
\Leftrightarrow{}& \disS p(\sVecNew,\yVecN\,|\,\Theta)   &>&   p(\sVecOld,\yVecN\,|\,\Theta)
\end{array}
$\\
\BOX\\[2mm]

The criterion (\ref{EqnCriterion}) can not be derived for arbitrary functions $f(\sVec,\yVecN)$ nor does it become obvious by considering
the original definition of $\FF(\KK,\Theta)$ in Eqn.\,\ref{EqnFreeEnergyTVEM}. Only thanks to the simplified form of $\FF(\KK,\Theta)$ derived
for Prop.\,3, the proof of Prop.\,6 becomes a straight-forward derivation (and the proof could be regarded as a formal, technical verification of
what may have been seen directly by considering the specific functional form of Eqn.\,\ref{EqnTruncatedF}).

By virtue of Prop.\,6, the free energy increase for the inner loop of Alg.\,1 can be ensured  if we, for a new state $\sVecNew$, find one state
in $\KKn$ with a lower joint probability than for $\sVecNew$. Also large numbers of new states, generated in parallel, can be compared in a bunch
to the states in $\KKn$. The set of states that increases the free energy most, can then be obtained through efficient partial sorting
\citep[e.g.][]{BlumEtAl1973}. Alternatively, efficient data structures such as heaps or soft heaps \citep[][]{Chazelle2000} to store the
%heaps \citep[][]{} or soft-heaps \citep[][]{} to store the
states of $\KKn$ according to their joint probabilities can be used. Any new set of states can then efficiently be compared with the
states in $\KKn$ using inequality (\ref{EqnCriterion}).

Given efficient updates of sets $\KKn$ using Prop.\,6 and the methods discussed above, the efficiency of the whole TV-E-step remains
to depend on efficiently finding new states $\KKtilde$ that are indeed sufficiently effective in increasing the free energy. 

%
%verification of free energy increase using Prop.\,6
%
%Efficient comparison of new states and efficient updates of $\KKn$ are, however, not sufficient to warrant an efficient increase of the
%free energy in Alg.\,1. The whole procedure will heavily depend on how likely it is that newly generated samples will indeed increase the free energy. Prop.\,6 provides an easy and efficiently computable rejection/%acceptance criterion for new samples to be included in $\KKn$. But if the likelihood of acceptance is too low, the whole variational optimization loop of Alg.\,1 will become inefficient. Let us therefore consider %different options to generate samples for Alg.\,1 in the following.
%
%
%Before we discuss different variation procedures further below, let us summarize the (partial) TV-EM procedure as a meta-algorithm (see Alg.\,1).\\[-2mm]
%
%% %
% \begin{eqnarray}
% %
% \qn(\sVec;\KK,\Theta) &=& \frac{\disS\phantom{\int}p(\sVec,\yVecN\,|\,\Theta)\phantom{\int}}
% {\disS\sum_{\sVecT\in\KKn}p(\sVecT,\yVecN\,|\,\Theta)}\,\delta(\sVec\in\KKn)\\
% %
% \FFHatHat(\KK,\Theta) &=& \sum_{n}\ \log\Big(\sum_{\sVec\in\KKn}\ p(\sVec,\yVecN\,|\,\Theta)\Big)
% %
% \end{eqnarray}
% %
%
\ \\
\noindent{\bf Blind search.} The easiest way to suggest new states $\KKtilde$ in Alg.\,1 is blind search. Such a search could be
realized by randomly (e.g.\,uniformly) sampling new states of the latent state, and then to use Prop.\,6 to compare these sampled
states to those in $\KKold$. Alternatively, one could use random (blind) variations of the old states to generate new states. Applying
Prop.\,6 would then realize a basic stochastic gradient ascent procedure which improves the free energy.
However, especially for large hidden spaces, the probability of newly generated states to increase the free energy 
will be small -- any blind search procedure will thus presumably be inefficient in general.

\ \\
\noindent{\bf Deterministic construction.} Instead of blindly and randomly searching new sets $\KK$ in Alg.\,1, an alternative would be
to deterministically construct newly suggested sets $\KKtilde$. Such a construction could use procedures already developed previously \citep[][]{LuckeEggert2010,SheltonEtAl2011,DaiEtAl2013,DaiLucke2014},
and Alg.\,1 would combine these constructions with the theoretical results derived for TV-EM.
For instance, most previous work used oracle (or selection) functions to construct sets $\KKtilde$.
%\citet[][]{SheltonEtAl2011,SheikhEtAl2014..}  used oracle (or selection) functions to construct sets $\KKtilde$.
A selection function could take the form of a scalar product \citep{SheltonEtAl2011}, of approximations or upper-bounds of marginal probabilities \citep[][]{LuckeEggert2010}, it could
be hand-crafted for specific (possibly relatively complex) generative models \citep[][]{DaiEtAl2013,DaiLucke2014}, or a selection function which is itself learned from data could be used \citep[][]{SheltonEtAl2017}.
Typically, a selection function is first used to determine for each data point $\yVecN$ a set $I^{(n)}$ of the most relevant latent variables (the
other variables are assumed with $s_h=0$ to not contribute to the generation of $\yVecN$). Given the set $I^{(n)}$, the set of states $\KKtilde^{(n)}$ can then
be constructed, for instance, by assuming a sparse combinatorics of the relevant latent variables:
\begin{equation}
\label{EqnKKtildeN}
\disT\KKtilde^{(n)}\, =\, 
\{\sVec\ |\ \sum_h{}s_h\leq{}\gamma\ \mbox{and}\ \forall{}h \not\in{}I^{(n)}: s_h=0 \}\,,
\end{equation}
where $\gamma$ parameterizes the considered sparsity level. For more details and for a  
visualization of such a constructed set $\KKtilde^{(n)}$ see, e.g., \citep[][Fig.\,2]{LuckeEggert2010} or \citep[][]{SheikhEtAl2014}.
The sets $\KKtilde^{(n)}$ were then directly used for the estimation of expectation values (\ref{EqnSuffStatMain}).

Instead of this previous direct use of $\KKtilde^{(n)}$ to define truncated posteriors, we can (based on the results obtained here)
use $\KKtilde^{(n)}$ as newly suggested states of Alg.\,1, and combine its elements with the old states $\KKoldsub^{(n)}$ to maximally increase the
free energy. Such a procedure would profit from well investigated methods to construct sets $\KKtilde$ for the different generative
models of previous work. On the downside, any new generative model would require the definition of new construction procedures, and such constructions
often make use of {\em ad hoc} assumptions such as sparsity \citep[][]{HennigesEtAl2010,DaiEtAl2013,HennigesEtAl2014}.
On the other hand, novel approaches to automate the construction of sets $\KKtilde$ \citep[see][]{SheltonEtAl2017} can directly be
applied in this context. In any case, deterministic construction closely links TV-EM to a series of previous pre-selection based
EM approximations \citep[][]{LuckeEggert2010} which have motivated this work initially. Importantly, these previous approaches can now be interpreted
as TV-EM with an {\em estimated} and one-step partial TV-E-step. `Estimated' because any direct definition of the states to define an approximation
does not guarantee the free energy to increase; and `one-step' because the variational loop of Alg.\,1 is replaced in the previous procedures
by one step which constructs the variational states.
%
%  the previous procedures are `one-step' because the variational loop of Alg.\,1 is replaced
%by one construction step.
%
%\ \\

\noindent{\bf Combination with sampling.} A further alternative to blind search would be a stochastic search for new states $\KKtilde$ in Alg.\,1
using knowledge given by the generative model. Already using the prior distribution of the generative model under consideration would be
more efficient than to blindly sample hidden states, e.g., using a uniform distribution. Samples from the prior would lie in areas of the
latent space where at least the average over all posteriors has a high probability mass. The space of high prior mass can still be very large,
however, and for any given data point, large posterior mass may be located in areas of the latent space very different from areas of high {\em average} posterior mass.

% such that new states that are likely to increase the free energy will again finally only be generated with relatively low probability.
Procedures that generate $\KKtilde$ by sampling new variational states in a data-driven way promise to be much more efficient. For instance,
if samples for newly suggested $\KKtilde^{(n)}$ are drawn from the posterior distribution $p(\sVec\,|\,\yVecN,\Theta)$, their joint probabilities
can be expected to be relatively high (and with them the free energy $\FFHatHat(\KK,\Theta)$). As the relative values of the joints are the crucial criterion
to find good sets $\KKtilde^{(n)}$, the common normalizer $p(\yVec\,|\,\Theta)$ is not relevant.
This observation together with the existence of a well established research field on efficient procedures to generate posterior samples,
would represent the advantages of such an approach. Potential disadvantages are that the highest posterior states will quickly be represented
by the sets $\KKn$ and that many new samples of $p(\sVec\,|\,\yVec,\Theta)$ may therefore already be contained in $\KK$.
Furthermore, posterior sampling is known to be challenging in high-dimensional latent spaces especially for discrete latents. Many of these
challenges may potentially carry over to TV-EM if samples from the posterior are used.

Preliminary work by \citet[][]{LuckeEtAl2018} investigates such sampling procedures using TV-EM for two concrete models: Binary Sparse Coding
\citep[][]{HaftEtAl2004,HennigesEtAl2010} and sigmoid belief networks \citep[][]{SaulEtAl1996,JordanEtAl1999}.
The approach emphasizes scalability and autonomous `black-box' optimization procedures. Both is achieved using a combination
of prior sampling and approximate marginal sampling, as both these sampling procedures can be defined without additional derivations. 
%
%
%Recent work by \citet[][]{LuckeEtAl2017} investigates the combination of TV-EM and sampling to train two generative models:
%binary sparse coding \citep[][]{HaftEtAl2014,HennigesEtAl2010} and (deep) sigmoid belief networks \citep[][]{SaulEtAl1996,JordanEtAl1999}. %, and shows the feasibility of the general procedure. 
%Scalability and autonomous `black-box' optimization procedures are in these applications of TV-EM to concrete generative models obtained
%using a combination of prior sampling and approximate marginal sampling. Efficiency is then further increase by facilitating marginal
%sampling using deep neural network approximations.
%
%
%
%
%
%
%
\subsection{TV-EM for Mixture Models}
\label{SecMixtureModels}
Mixture models can be regarded as complementary to multiple-cause generative models. In their different versions, they are
among the most widely applied generative data models and very successful in many tasks of image or sound processing as well as
for general pattern analysis tasks \citep[e.g.][]{McLachlanPeel2004,Duda2007,ZoranWeiss2011,PoveyEtAl2011}. In contrast to multiple-cause models,
any observed variable $\yVec$ is in mixture models assumed to be generated by exactly one cause, i.e., one class. 
In their most standard version, a discrete hidden variable $c$ is taken to represent one of $C$ causes or {\em clusters} and to generate data 
via a noise distribution $p(\yVec\,|\,c,\Theta)$. The data distribution assumed by a mixture model is thus given by:
\begin{equation}
\label{EqnMixtureModel}
p(\yVec\,|\,\Theta) \ =\ \sum_{c=1}^{C} \pi_c p(\yVec\,|\,c,\Theta)\ \ \ \mbox{with}\ \ \sum_{c=1}^{C}\pi_c=1,
\end{equation}
where the prior parameters $\pi_c=p(c\,|\,\Theta)\in[0,1]$ are model parameters commonly referred to as mixing proportions.

As the hidden variable is discrete, TV-EM can directly be applied. We here only change the notation slightly to be more consistent with
the conventional mixture model notation, i.e., we replace $\sVec$ for the hidden variable by the integer $c\in\{1,\ldots,C\}$. 
The sets $\KKn$ then consequently contain subsets of class indices, $\KKn\subseteq\{1,\ldots,C\}$. %Alternatively, we could have maintained
%the original notation by using states $\sVec$ with just one non-zero entry at position $c$ (which is sometimes referred to as `one-hot-coding').

Mixture models may not be considered as the typical application domain of variational EM procedures, but they were
used as example applications already early on: \citet[][]{NealHinton1998}, for instance, 
motivated the application of variational EM by its increased efficiency for mixture models,
and we will see below that the same motivation applies for the application of TV-EM.

For this example we consider TV-EM with a full E-step (Eqns.\,\ref{EqnTVEStep} to \ref{EqnTVThird}).
Based on Eqn.\,\ref{EqnTVEStep}, the TV-E-step for mixture models corresponds to finding those states $c$ that
globally maximize the simplified free energy $\FF(\KK,\Theta)$ (\ref{EqnTruncatedF}) of Prop.\,3 w.r.t.\,$\KK$.
Let us again constrain all sets $\KKn$ to be of the same size, i.e., $|\KKn|=C'\leq{}C$ in this case. 
Because of the form of the free energy in Eqn.\,\ref{EqnTruncatedF}, we can now show that it is sufficient and efficient to pair-wise
compare all joint probabilities in order to find optimal $\KKn$.

\ \\
{\bf Proposition 7}\\*
Consider the application of TV-EM to a mixture model given by (\ref{EqnMixtureModel}) with $C$ clusters, % and TV-EM , its truncated free energy of Eqn.\,\ref{} (with $\sVec$ replaced by $c$),
and let $\KKn$ (with $|\KKn|=C'$) be the set of variational states for a data point $\yVecN$.  

Then the free energy is maximized in the TV-E-step (\ref{EqnTVEStep}) if for all $n=1,\ldots,N$ the sets $\KKn$ contain the $C'$ clusters with the largest joint probabilities, i.e., if
\begin{eqnarray}
\label{EqnLargestJoints}
%
%\mbox{for all}\ \ n=1,\ldots,N\ \ 
\mbox{for all}\ \ c\in\KKn\ \mbox{ and }\ c'\not\in\KKn:\ \ p(c,\yVecN\,|\,\Theta) \geq p(c',\yVecN\,|\,\Theta).
\end{eqnarray}
Such a maximum can be found using $\calO(NC)$ comparisons of joint probabilities.

\ \\
\noindent{}{\bf Proof}\\[1mm]
Let us consider the set $\KK=(\KK^{(1)},\ldots,\KK^{(N)})$ for which each $\KKn$ fulfills criterion (\ref{EqnLargestJoints}).
If we now replace for a specific but arbitrary $n$ an arbitrary $c\in\KKn$ by a $c'\not\in\KKn$ then by our definition of $\KK$:
$
\begin{array}{lrcl}
p(c',\yVecN\,|\,\Theta) \leq p(c,\yVecN\,|\,\Theta).
\end{array}
$
According to Prop.\,6, the free energy is then decreased or remains constant. As for any $n$ a change of any cluster $c\in\KKn$ results in a
decreased or constant free energy, the set $\KK$ must represent a global maximum of $\FF(\KK,\Theta)$ (no better set can be found).\\[2mm]
Regarding the complexity of finding the maximum, let ${\cal C}^{(n)}$ be a list of all joint
probabilities $p(c,\yVecN\,|\,\Theta)$ for a fixed data point $\yVecN$. All such lists ${\cal C}^{(n)}$ are of size $C$.
Let us first suppose that all elements in ${\cal C}^{(n)}$ have different values.
Finding the set $\KKn$ which fulfills (\ref{EqnLargestJoints}) is then the problem of finding the $|\KKn|$ largest elements in a list of $C$ elements.
This partial sorting problem is according to \citep[][]{BlumEtAl1973} solvable using $\calO(C)$ comparisons of the elements. In case of two or more identical
elements in ${\cal C}^{(n)}$, the same partial sorting procedure returns a list (i.e., a set $\KKn$) which also fulfills $(\ref{EqnLargestJoints})$ (but
there may now be more than one such $\KKn$ satisfying the criterion). By repeating the procedure $N$ times (once for each data point $n$), we can define 
a set $\KK=(\KK^{(1)},\ldots,\KK^{(N)})$ for which each $\KKn$ fulfills (\ref{EqnLargestJoints}). The set $\KK$ is therefore (A)~computable using $\calO(NC)$ comparisons of joint probabilities,
and (B)~it maximizes the free energy because all its $\KKn$ satisfy $(\ref{EqnLargestJoints})$.\\
\BOX\\[2mm]
Prop.\,7 defines a concrete deterministic and efficient procedure applicable to any mixture model of the form (\ref{EqnMixtureModel}). In practice, the procedure requires to actually
compute all the joint probabilities first (at least up to a common factor) in order to realize the required comparison. The computational demand for computing the joints is $\calO(NC)$ times
the computations required for the evaluation of each joint (which is usually proportional to the number data space dimensions $D$), e.g., $\calO(NCD)$ for Gaussian Mixture Models (GMMs).

As mixture models have latent state spaces of size linear in $C$, it may not be considered surprising
that TV-EM is applicable using $\calO(NC)$ comparisons of joint probabilities. After all, an exact E-step of standard EM, e.g., for GMMs, also only requires $\calO(NCD)$ computations.
However, TV-EM can reduce the required computations for the M-step because it uses exact zeros, i.e., M-steps with expectation values (\ref{EqnSuffStatMain}) can be shown to be less complex.
The price to pay for this reduction is that the formal optimization problem of the TV-E-step (Eqn.\,\ref{EqnTVEStep}) is defined on a space larger than the original state space. For the here (and throughout the paper) assumed equally sized sets $\KKn$ (here $|\KKn|=C'$), the number of all possible subsets $\KKn\subseteq\{1,\ldots,C\}$ is $\left(\begin{array}{c}C\\C'\end{array}\right)$. Prop.\,7 ensures that we do not have to exhaustively visit all these subsets but can find the maximizing set for each $n$ efficiently. Again, this efficiency result is ultimately due to the simplified form of the free energy (\ref{EqnTruncatedF}).

\begin{algorithm}[t]\vspace{1.5mm}
%\dontprintsemicolon
init model parameters $\ThetaOld$;\algBreak
% init variational states $\KKold$;\algBreak
%
\Repeat{parameters $\Theta$ have sufficiently converged\vspace{1.5mm}}
{
%set $\ThetaHat=\Theta$\\%\algBreak
%
  \For{$n=1:N$}{
    \For{$c=1:C$}{
 	compute $p(c,\yVecN\,|\,\ThetaOld)$;
    }
    define $\KKn$ to contain the $C'$ indices $c$ with largest $p(c,\yVecN\,|\,\ThetaOld)$;  
  }
compute $\ThetaNew$ that maximizes $\disS Q(\Theta)=\sum_{n=1}^{N}\E{\log\big( p(c,\yVecN\,|\,\Theta)\big)}_{\qn(c;\KK,\ThetaOld)}$;\\
$\ThetaOld = \ThetaNew$;\algBreak
}
% $
% \begin{array}{rcl}
% \qn(\sVec;\KK,\Theta) &=& \frac{\disT\phantom{\int}p(\sVec,\yVecN\,|\,\Theta)\phantom{\int}}
% {\disT\sum_{\sVecT\in\KKn}p(\sVecT,\yVecN\,|\,\Theta)}\,\delta(\sVec\in\KKn)\\[1mm]
% %
% \FFHatHat(\KK,\Theta) &=& \sum_{n}\ \log\big(\sum_{\sVec\in\KKn}\ p(\sVec,\yVecN\,|\,\Theta)\big)
% %
% \end{array}
% $
%
\caption{Application to Mixture Models.\label{AlgTwo}}
\end{algorithm}

The application of TV-EM to mixture models (as summarized by Alg.\,2) now allows for interpreting earlier applications to mixture models within the derived free energy framework. 
Truncated approximations were previously applied to mixture models, e.g., to GMMs in work by \citet[][]{SheltonEtAl2014} and later by \citet[][]{HughesSudderth2016}.
The cluster finding procedure used by \citet[][]{SheltonEtAl2014} can in the light of this study be recognized as an estimated TV-E-step (using Gaussian Processes to construct the sets $\KKn$),
while the constrained likelihood optimization for exponential family mixtures as used by \citet[][]{HughesSudderth2016} can be recognized as a full TV-E-step.
%Also the E-steps of more intricate mixture models which include latents for translation invariance \citep[][]{DaiLucke2014} can be recognized as estimated TV-E-steps.
%
For Poisson mixtures, \citet[][]{ForsterLucke2017} directly applied the TV-EM algorithm (Alg.\,2) suggested by Prop.\,7. For all the above applications, our theoretical results show that
the free energy (\ref{EqnTruncatedF}) is the underlying objective function which is maximized. Furthermore, for the algorithms by \citet[][]{HughesSudderth2016}, \citet{ForsterLucke2017} and
\citet{ForsterLucke2018} the TV-EM application to mixture models warrants that the free energy is provably monotonically increased. Monotonic increase follows from Prop.\,5
and has not been shown previously. Furthermore, our results apply for {\em any} mixture model of the form (\ref{EqnMixtureModel}) and for Alg.\,2 as well as for corresponding
partial EM versions.

The main motivation and focus of the previous truncated approximations for mixture models \citep[][]{HughesSudderth2016,ForsterLucke2017,ForsterEtAl2018} was an increase in efficiency.
The source for the reduction of computational efforts were hereby the `hard' zeros introduced by truncated posteriors, which significantly reduced the required number of
numerical operations in the M-step. The work by \citet[][]{HughesSudderth2016} focuses on this M-step complexity reduction, and they empirically find that the whole EM
optimization procedure only required about half the operations compared to exact EM. Also, \citet[][]{ForsterLucke2017} focus on the complexity reduction provided by the M-step
and observe a similar efficiency increase for Poisson mixtures. Notably, TV-EM does not negatively effect the final likelihood values that were
reported by these studies. On the contrary, faster convergence and higher final likelihoods for different datasets were observed
empirically \citep[][]{HughesSudderth2016,ForsterEtAl2018,LuckeForster2019}. This is due to TV-EM avoiding local likelihood optima more effectively than exact EM -- an effect
that has also been observed for sparse coding models and (preselection-based) truncated approximations \citep[][]{ExarchakisEtAl2012}.

Except of reducing the M-step complexity by TV-EM \citep[][]{HughesSudderth2016,ForsterLucke2017}, the here derived results point to a further possibility for complexity reduction.
In deriving partial TV-EM (Eqns.\,\ref{EqnTVEMOptStepsParialFinal}) we have shown that the free energy (\ref{EqnTruncatedF}) also monotonically increases for partial TV-E-steps.
A full maximization is, hence, not required to obtain an algorithm that provably increases (\ref{EqnTruncatedF}). As efficient criteria to verify increased free energies are available,
very efficient partial E-steps were investigated in work parallel to this study: \citet[][]{ForsterLucke2018} have thus shown that clustering algorithms in which each EM iteration scales
sublinearly with $C$ can be derived. Numerical experiments then show that the free energy (and likelihood) objective is still efficiently increased, which provides evidence for clustering
being scalable sublinearly with $C$ \citep[for details see][]{ForsterLucke2018}.
%
%Empirical studies have so far been conducted using standard and some more intricate forms of mixture models. Any theoretical guarantees here provided for such models do directly carry over to any mixture model, however. 
%
%
%
%
%
%
\subsection{TV-EM and `Hard EM'}
\label{SecHardEM}
%
%`Hard' EM is a widely used version of EM which replaces exact EM by 
%
A very wide-spread approach to optimize parameters of a given generative model is `hard EM' also known as zero-temperature EM, MAP approximation, Viterbi training or classification EM.
As the name suggests, `hard EM' is typically introduced as an EM-like algorithm in which the computation of the full posterior in the E-step is replaced by the computation of the
state $\sVec$ with maximum a-posterior (MAP) probability.\ \,In the M-step, the model parameters are then updated by only\hspace{90mm}\vspace{2mm}
\SetAlCapHSkip{0.2em}
\DecMargin{1.0em}
\begin{wrapfigure}[16]{r}{0.66\textwidth}
\begin{minipage}{0.65\textwidth}
\ \\[-12mm]
%\myvanish{
\begin{algorithm}[H]
%\dontprintsemicolon
init model parameters $\ThetaOld$;\\ %\algBreak
% init variational states $\KKold$;\algBreak
%
\Repeat{parameters $\Theta$ have sufficiently converged\vspace{1.5mm}}
{
%set $\ThetaHat=\Theta$\\%\algBreak
%
  \For{$n=1:N$}{
     $\sVec^{(n)} = \hspace{1mm}\argmax{\tilde{\sVec}}\big\{p(\tilde{\sVec}\,|\,\yVecN,\ThetaOld)\big\}$;\phantom{iiiiiiii}({\bf hard E-step})
%    define $\KKn$ to contain the $C'$ indices $c$ with largest $p(c,\yVecN\,|\,\ThetaOld)$;  
  }
  $\ThetaNew  =\hspace{1mm}\argmax{\Theta}\big\{\sum_{n}{}\log\big(p(\sVecN,\yVecN\,|\,\Theta)\big)\big\}$;\phantom{m}({\bf M-step})\algBreak
  $\ThetaOld = \ThetaNew$;\vspace{2mm}
}
% $
% \begin{array}{rcl}
% \qn(\sVec;\KK,\Theta) &=& \frac{\disT\phantom{\int}p(\sVec,\yVecN\,|\,\Theta)\phantom{\int}}
% {\disT\sum_{\sVecT\in\KKn}p(\sVecT,\yVecN\,|\,\Theta)}\,\delta(\sVec\in\KKn)\\[1mm]
% %
% \FFHatHat(\KK,\Theta) &=& \sum_{n}\ \log\big(\sum_{\sVec\in\KKn}\ p(\sVec,\yVecN\,|\,\Theta)\big)
% %
% \end{array}
% $
%
\caption{The `Hard EM' algorithm.\label{AlgHardEM}}\vspace{2.5mm}
\end{algorithm}
\end{minipage}
\end{wrapfigure}

\ \\[-17mm]
\noindent{}considering this maximum {\em a-posterior} state. Alg.\,\ref{AlgHardEM} shows a standard form of the `hard EM' algorithm. `Hard EM' is often regarded as an {\em ad hoc} procedure, which replaces a computationally intractable full posterior in the E-step by a maximization. Such a maximization is easier because it can be reformulated as the maximization of a computationally tractable objective, the joint $p(\sVec,\yVecN\,|\,\Theta)$ (as the normalizer $p(\yVecN\,|\,\Theta)$ does not depend on
$\sVec$). `Hard EM' can, however, also be derived from annealed versions of EM. 
%more systematically, and the presumably most common derivation is to consider annealed EM.
Annealed EM is a procedure usually introduced in order to avoid local optima \citep[e.g.][]{UedaNakano1998,Sahani1999}. A temperature parameter is introduced which forces the probability values
of the posterior to become more equal. %While annealed EM algorithms are obtained for high temperatures, `hard EM' is obtained if instead the limit to zero temperature is considered (see Appendix for details).
%
%\citep[compare, e.g.,][for Hidden Markov Models]{CohenSmith2010,AllahverdyanGalstyan2011,TurnerSahani2011}.
%
%In any case, `hard EM' remains rather an heuristic approach; it requires to take a limit (temperature zero), and it implicitly assumes that states that maximize the posteriors can actually be found. 

By considering Alg.\,\ref{AlgHardEM}, `hard EM' can be formulated by replacing the posterior by a $\delta$-function centered at the MAP state (often termed Dirac-$\delta$ in the continuous and
Kronecker-$\delta$ in the discrete case). $\delta$-functions have, frequently, merely been considered in this context as a way to explicitly formulate the function which replaces the
full posterior in the MAP approximation. However, the link between MAP approximation and variational EM has also early on been pointed out \citep[][]{JordanEtAl1997}.
Following the introduction of truncated posteriors for TV-EM, and by virtue of Props.\,1 and 2, we will here treat the $\delta$-functions within the variational framework of TV-EM.
For this, we consider the states for which the $\delta$-functions are non-zero as variational parameters of truncated distributions. Such a formulation then corresponds
to a TV-EM algorithm with sets $\KKn$ which each contain just one element, i.e.\ $\KKn=\{\sVecN\}$ for all $n$. More precisely, we can for this boundary case of TV-EM show the following:

%
%identify the objective optimized by `hard' EM as a free energy that is naturally obtained without the requirement to introduce annealing temperatures
%and limits thereof. For this consider TV-EM with sets of variational states $\KKn$ which each contain just one element, i.e. $\KKn=\{\sVecN\}$ for all $n$. We then obtain: %For this special case of TV-EM we then obtain: 
%
\ \\
{\bf Proposition 8}\\*
Consider a generative model $p(\sVec,\yVec\,|\,\Theta)$ with discrete latents $\sVec$.
Then `hard EM' for this model (Alg.\,\ref{AlgHardEM}) is equivalent to a TV-EM algorithm which uses sets $\KKn$ with just one element each.\\[2mm]
{\bf Proof}\\*[1mm]
Note that all results derived for truncated variational distributions, so far, apply for arbitrary (non-empty) sets $\KKn$.
For a TV-EM algorithm with sets $\KKn$ that contain just one element each, we can denote these elements by $\sVecN$, i.e., $\KKn=\{\sVecN\}$.
Let us first consider the simplified truncated free energy (\ref{EqnTruncatedF}) derived in Prop.\,3. For $\KKn=\{\sVecN\}$ we then obtain:
\begin{equation}
\FF(\KK,\ThetaOld) = \disS\sum_{n=1}^{N}\log\big(\backHalf\sum_{\ \sVec\in\KKn} p(\sVec,\yVecN\,|\,\ThetaOld)\,\big) = \disS\sum_{n=1}^{N}\log\big( p(\sVecN,\yVecN\,|\,\ThetaOld)\,\big)\,.
%
%                \ =\  \disS\log\Big( \prod_{n=1}^{N} p(\sVecN,\yVecN\,|\,\ThetaOld)\Big)
%
\end{equation}
The maximum of $\FF(\KK,\ThetaOld)$ w.r.t.\ $\KK=(\KK^{(1)},\ldots,\KK^{(N)})$ can be found by individually maximizing each summand $\log\big( p(\sVecN,\yVecN\,|\,\ThetaOld) \big)$ w.r.t.\ $\sVecN$.
The states $\sVecN$ that maximize the summands are then the same as those computed in the `hard' E-step of Alg.\,\ref{AlgHardEM} because:
\begin{eqnarray}
\sVecN &=& \argmax{\sVec}\big\{ \log\big( p(\sVec,\yVecN\,|\,\ThetaOld) \big\}\ =\ \argmax{\sVec}\big\{ p(\sVec\,|\,\yVecN,\ThetaOld) \big\}\,.
\end{eqnarray}
The new set $\KKnew$ which is computed by the TV-E-step (\ref{EqnTVEStep}) is consequently given by $\KKnew=(\{\sVec^{(1)}\},\ldots,\{\sVec^{(N)}\})$.
% Hence, the maximization of the free energy w.r.t.\ the sets $\KKn$ required by the TV-E-step (...), becomes equivalent to individually maximizing the joints w.r.t.\ the states $\sVec$, i.e.,
%
%\begin{eqnarray}
%
%\KKnew=(\{\sVec^{(1)}\},\ldots,\{\sVec^{(N)}\}) \phantom{iiiiii}\mbox{where for all $n$ applies}\phantom{iii} \sVecN = \argmax{\tilde{\sVec}}\{  p(\tilde{\sVec},\yVecN\,|\,\Theta)  \}.
%
%\end{eqnarray}
%
In the TV-M-step (\ref{EqnTVMStep}) the truncated free energy $\FF(\KKnew,\ThetaOld,\Theta)$ is then optimized w.r.t.\ $\Theta$. 
When $\KKn=\{\sVecN\}$, a truncated distribution $\qn(\sVec;\KKnew,\ThetaOld)$ is unequal zero only for the state $\sVec=\sVecN$, i.e., $\qn(\sVec;\KKnew,\ThetaOld)=\delta(\sVec=\sVecN)$.
As, additionally, the entropy term of the free energy vanishes for such $\qn$, the free energy $\FF(\KK,\ThetaOld,\Theta)$ reduces to: % the same form as the simplified free energy:
\begin{eqnarray}
  \FF(\KKnew,\ThetaOld,\Theta) &=&
    \sum_{n=1}^{N} \bigg[
      \sum_{\sVec}\ \qn(\sVec;\KKnew,\ThetaOld)\  
       \log\left( p(\sVec,\yVecN\,|\,\Theta)\right) 
      \bigg] 
      + H(q(\sVec;\KKnew,\ThetaOld))\nonumber\\
                            &=&
    \sum_{n=1}^{N} \bigg[
      \sum_{\sVec}\ \delta(\sVec=\sVecN)\  
       \log\left( p(\sVec,\yVecN\,|\,\Theta)\right)
                            \ =\ 
      \sum_{n=1}^{N} \log\Big( p(\sVecN,\yVecN\,|\,\Theta)\Big).\nonumber
\end{eqnarray}
Maximization of the free energy $\FF(\KK,\ThetaOld,\Theta)$ w.r.t.\,$\Theta$ is thus equivalent to the `hard' M-step of Alg.\,\ref{AlgHardEM}. As the $\sVecN$ used in the hard M-step are
precisely those computed in the `hard' E-step, TV-EM with just one element for each $\KKn$ is equivalent to `hard EM' (Alg.\,\ref{AlgHardEM}).\\
\BOX\\
\ \\
In practice, the maximization in the hard E-step is often difficult to accomplish, such that states $\sVec$ are computed that only approximately maximize the
posteriors $p(\sVec\,|\,\yVecN,\Theta)$. Such a partial `hard EM' approach can then be shown to correspond to TV-EM with a partial E-step (Eqns.\,\ref{EqnTVEMOptStepsParialFinal}),
and the proof follows along the same line as the proof for Prop.\,8.
%
%
%This observation is also relevant for the common practice in using the 
%previous best or approximately best states $\sVecN$ as starting points for the next optimization within a hard EM step. Considering ... such a practice warrants
%that the fr
%

The free energy for `hard EM' as derived in the proof of Prop.\,8 can be rewritten by replacing $\KK=(\{\sVec^{(1)}\},\ldots,\{\sVec^{(N)}\})$ with $\sVec^{(1:N)}$:
\begin{eqnarray}
\label{EqnFreeEnergyHard}
 \FF(\sVec^{\,(1:N)},\Theta) =  \sum_{n=1}^{N}\log\Big( p(\sVecN,\yVecN\,|\,\Theta)\Big)    \leq \LL(\Theta).
\end{eqnarray}
Note that objective functions such as \refp{EqnFreeEnergyHard} (or very similar ones)
have already previously been discussed. For instance, the `hard EM' objectives as stated for
generative models such as Gaussian Mixtures or Hidden Markov Models \citep[e.g.][]{JuangRabiner1990,CeleuxGovaert1992,CohenSmith2010} directly relate to the `hard EM'
free energy (\ref{EqnFreeEnergyHard}). Such objectives were usually defined as a tool to interpret the heuristically introduced `hard EM' algorithm but
work, e.g., by \citet[][]{JordanEtAl1997}, early on pointed out that `hard EM' for Hidden Markov Models can be understood as a variational approach. 
Here, we obtain `hard EM' as a special case of TV-EM. This means that TV-EM represents a natural interpolation between full EM (with sets $\KKn=\Omega$) and
`hard EM' (with sets $\KKn=\{\sVecN\}$). By choosing intermediate
sets $\KKn$ (with $1<|\KKn|<|\Omega|$), TV-EM can consequently trade-off accuracy vs.\ efficiency. In a number empirical studies, such interpolations have been found
to work very well in practice. For instance, \citet[][]{ForsterEtAl2018} found $|\KKn|=15$ non-zero states to work best for Poisson mixtures with $10\,000$
classes applied to MNIST, \citet[][]{HughesSudderth2016} found truncated distributions with few (but more than one) non-zero states to work well for topic models,
and \citet[][]{HughesSudderth2016,LuckeForster2019} found few non-zero states to work best for standard Gaussian Mixture Models.
Also previously, with truncated approximations using estimated E-steps, sets $\KKn$ with few states (but more than one) were found to work well.
Such approaches have been applied to invariant latent variable models \citep[][]{DaiLucke2012b,DaiEtAl2013}, invariant mixture models \citep[][]{DaiLucke2014},
as well as standard Gaussian mixtures \citep[][]{SheltonEtAl2017}.

Notably, all such `almost hard EM' approaches are non-trivial generalization of `hard EM'. For `hard EM' we do not
necessarily require Prop.\,5 in order to show that the free energy monotonically increases. This is because the
free energy and its simplified version given by Prop.\,3 coincide for one element per $\KKn$. In this case TV-EM
(i.e., `hard EM') becomes a straight-forward coordinate-wise ascent approach w.r.t.\ this objective.
However, any generalization to more than one state with non-zero probability
changes the free energy objective $\FF(\KK,\Theta)$ to the more general form (\ref{EqnTruncatedF}) given by Prop.\,3. Taking derivatives of $\FF(\KK,\Theta)$
in (\ref{EqnTruncatedF}) w.r.t.\,$\Theta$ is now different from taking derivatives of the free energy $\FF(\KK,\ThetaOld,\Theta)$ in (\ref{EqnFreeEnergyTVEMOrg}).
If the standard M-step equations for a given generative model are maintained (which optimize $\FF(\KK,\ThetaOld,\Theta)$) then we require Prop.\,5 to show that
TV-EM indeed monotonically increases the free energy.
%
%Note that annealed EM (see Appendix) does not provide such an embedding. 
%In principle
%it would be possible to consider annealed posteriors as variational distributions but then it would be required to also consider the parameters of
%the annealed posteriors as variational parameters (in order to guarantee that standard M-step updates can be maintained).
%Using a rigorous treatment similar to the one for truncated distributions considered in this work, results similar to Props.\,4 and 5 may be derivable,
%and for the limit to zero temperature, results similar to those of Props.\,1 and 2 would have to be derived.
%

Hence, while algorithmically Prop.\,8 does not add much novelty, the well known `hard EM' approach can now be understood as a boundary case of a range
of possible approximations provided by TV-EM. Furthermore, any results for TV-EM including partial E- or M-steps also apply for `hard EM'.
The free energy of `hard EM' and its properties can also be derived more directly, i.e., without considering `hard EM' as a special case of TV-EM. The objective function formulations mentioned
above \citep[e.g.][]{JuangRabiner1990,CeleuxGovaert1992,CohenSmith2010} and more explicitly \citet[][]{JordanEtAl1997} at least partly provide such relations. For completeness, we
state some properties of the free energy for `hard EM' in \ref{AppC} together with their derivations from TV-EM results.
\section{Discussion}
\label{SecDiscussion}
We have defined and analyzed a novel variational approximation of expectation maximization (EM). Our approach is based on truncated {\em a-posteriori} distributions with latent states as variational parameters.
%In this work we have studied the theoretical foundation of truncated variational approximations to the expectation maximization (EM) approach. 
%
Our first set of results (Props.\ 1 and 2) generalize the variational free energy approach as introduced, e.g.,\ by \citet{SaulEtAl1996,NealHinton1998,JordanEtAl1999} by including discrete variational distributions with exact zeros. While this generalization is required in order to study truncated variational distributions, Props.\,1 and 2 also apply for any other discrete distributions with exact (`hard') zeros, i.e., these results are not restricted to the specific truncated distributions of Eqn.\,\ref{EqnQMain}. 
%As such, Props.\ 1 and 2 generalize the standard text book derivation of variational free energies \citep[e.g.][]{Bishop2006,Murphy2012,Barber2012}, i.e., the standard additional demand of $q(\sVec)>0$ can be dropped for the discrete case.
Props.\ 3, 4, 5 and Corollary 1 then represent results specific to variational distributions in the form of truncated posteriors (Eqn.\,\ref{EqnQMain}).
%, including the concise form of truncated free energies and proofs of monotonic free energy increase.
Props.\ 6, 7 and~8, finally, represent example applications of the theoretical results.
\paragraph{Relation to Preselection-Based Truncated Approximations.}
Previous algorithms based on truncated approximations of expectation values \citep[][]{LuckeSahani2008,LuckeEggert2010,HennigesEtAl2014,DaiLucke2014,SheikhEtAl2014,SheikhLucke2016,SheltonEtAl2017} have
motivated this work. These contributions have directly approximated expectation values by exploiting sparsity of latent activities \citep[][]{LuckeSahani2008} and by additionally using a preselection procedure in what was termed Expectation Truncation \citep[ET;][]{LuckeEggert2010}. 
%
%Preselection in ET approximations meant the use of a fast estimation procedure to identify those hidden variables that could be expected to be relevant for a given data point $\yVecN$. Only based on those variables, sets $\KKn$ were constructed by combining the relevant states in specific ways. Such estimation procedures made use of so called {\em selection function} (or oracle functions) that gave high scores to latent variables relevant for a given data point $\yVecN$. 
%
Based on the theoretical framework derived in this work and Sec.\,\ref{SecSampling}, all previous selection-based approaches can be considered as approximations of TV-EM. {\em Expectation Truncation} can thus be embedded into the framework of variational approaches. While estimated E-steps using ET may potentially {\em decrease} the truncated free energy of Prop.\,3, the general effectiveness and efficiency of previous ET applications may be taken as evidence for the efficiency and effectiveness of truncated approximations in general. Such efficiency and effectiveness can now be generalized and theoretical guarantees for tractable free energies are available. Furthermore, and maybe most importantly, TV-EM can now provide (A)~a straight-forward generalization and applicability to very advanced (including deep) data models, and (B)~it avoids any additional effort to define model specific selection functions (compare discussion of `black box' procedures below).
%
%Maybe more signigicantly, the use of selection functions required to realize truncated approximations sets limits to the generative models that can be considered. More complex models \citep[e.g.][]{DaiLucke2014,DaiEtAl2013} thus required significant effort to define and tune selection functions, which motivated learnable such functions \citep[][]{SheltonEtAl2017}. Especially for tackling the challenge of training increasingly more advanced data models (including deep directed models) thus poses problems preselection-based approaches.

In one aspects, ET provides a result that has not been addressed here, however: it can be shown that preselection allows for the definition of smaller generative models defined per data point, and that optimizing parameters of such smaller models approximately optimizes the parameters of the original larger generative model. This result, valid for a large class of generative models (but not for all), has been used in select-and-sample approaches \citep[][]{SheltonEtAl2012,SheltonEtAl2015}, was formally proven in \citep[][]{SheikhLucke2016}, and carries over to recent work using selection functions that are themselves learned from data \citep[][]{SheltonEtAl2017}.
\paragraph{Relation to mean-field and `sparse EM'.} 

As TV-EM is a variational approximation for models with discrete latents, the main standard class of variational approaches for comparison
are factored variational (i.e.\,{\em mean-field}) approaches \citep[][]{SaulJordan1996,JordanEtAl1999}. Compared to fully factored
approaches (Eqn.\,\ref{EqnFreeEnergyFVEM}), a main difference to TV-EM is that truncated approaches do not assume posterior independence. Assuming independence (i.e., neglecting explaining-away effects) has been observed to negatively impact likelihood optimization for different types of generative models \citep[][]{IlinValpola2003,MacKay2003,TurnerSahani2011,SheikhEtAl2014}.
In order to address such potentially harmful consequences, partly factored approaches have been studied \citep[][]{SaulJordan1996,MacKay2003,Bouchard2009}. Deviations from fully factored approaches do,
however, often go along with increased analytical and computational effort, and may even provide only limited improvements compared to mean-field \citep[e.g.][for a discussion]{TurnerSahani2011}.
In comparison, TV-EM does not assume independence. Computational tractability is instead achieved by approximating posteriors on sufficiently small subsets of the latent space. 

Other than mean-field variational approximations, `sparse EM' is another (less frequently applied) alternative which was discussed early on in the very influential work by \citet[][]{NealHinton1998}. Notably, \citet[][]{NealHinton1998} never explicitly mentioned {\em factored} variational distributions in their work. Instead the paper discussed with `sparse EM' a variational approach that shares more similarities with TV-EM than with mean-field. In their `sparse algorithm', \citet[][]{NealHinton1998} suggested a variational distribution defined on a subset of the state space which only contains `plausible' values of the latents (their Sec.\,5). The probabilities outside of this set were `frozen' to values of an earlier iteration but updated once in a while during learning. The procedure there described is not efficient, e.g.\ for latent variable models with large state spaces, because `sparse EM' still requires an occasional evaluation of {\em all} latent states. As \citet[][]{NealHinton1998} discuss an application to a mixture model, this shortcoming is not very relevant for their paper. In contrast to `sparse EM', truncated approximations assume exact zeros and can thus realize approximations without ever having to evaluate all hidden states while they are still able to find subsets of the latent space with high posterior mass. % The explicite form of TV-EM (REF) may provide the most direct 
However, despite these differences, the results obtained for TV-EM in this work may be regarded as connecting back to an initial (and never followed up) train of thoughts expressed in the work by \citet[][]{NealHinton1998}.

\paragraph{Relation to `hard EM'.} A further, very influential class of approximate EM alogrithms is `hard EM' (alias `zero temperature EM', `classification EM' or EM using MAP approximations).
`Hard EM' approaches have been used for many types of data models including deep models, and they were often observed to work very well in practice. Because of their wide-spread use, e.g., in domains
such as sparse coding \citep[][]{OlshausenField1996,MairalEtAl2010}, compressive sensing \citep[][]{Donoho2006,Baraniuk2007} but also for relating generative modeling and deep
learning \citep[][]{PatelEtAl2016}, 'hard EM' may even be considered more wide-spread than any conventional or novel variational EM approaches. As shown in Sec.\,\ref{SecHardEM}, we here identified
`hard EM' as a TV-EM algorithm with sets of variational parameters $\KKn$ just containing one state (Prop.\,8). `Hard EM' (including versions with partial posterior optimization)
are consequently embedded into the variational EM framework and monotonically increasing free energies can be provided. 
While such results can also be derived directly \citep[compare, e.g.,][]{JuangRabiner1990,CeleuxGovaert1992,CohenSmith2010,JordanEtAl1997} TV-EM provides concrete procedures to generalize any `hard EM' algorithm to algorithms with multiple `winning' states.
%
%Previous applications \citep[e.g.][]{DaiEtAl2013,DaiLucke2014,SheltonEt2014} do not use the TV-EM framework derived here but their E-steps can be interpreted as estimated TV-E-steps. More recent contributions \citep[][]{HughesSudderth2016,LuckeForster2017,ForsterLucke2017} are, on the other hand, 
%directly based on or can be directly related to the TV-EM algorithms derived in this study.
%
%In the light of Prop.\,8, also all previous applications of preselection-based truncated EM \citep[][]{LuckeEggert2010,SheikhEtAl2014,HennigesEtAl2014,SheikhLucke2016} may be considered as generalized `hard' EM approaches. However, the number of states in $\KKn$ for these approaches is often considerably larger than just a few states.
%
%, were `almost hard EM' generalizations of `hard EM' have shown to work well in practice either by directly using the results of this study 
%\citep[][]{SheltonEt2014,HughesSudderth2016,LuckeForster2017}
%\citep[][]{SheltonEt2014,HughesSudderth2016,LuckeForster2017}, Poisson mixtures \citep[][]{ForsterLucke2017}

%In addition to such generalizations, the identification of `hard EM' as a variational EM procedure may have implications for many different types of algorithms. 
For instance, for training of deep networks with `hard EM' \citep[e.g.][]{PoonDomingos2011,OordEtAl2014,PatelEtAl2016}, generalizations with more than one non-zero state may be considered very interesting
\citep[especially considering the effectiveness of such approaches for mixture models; e.g.][]{HughesSudderth2016,ForsterLucke2017}. Similarly, time-series models such as Hidden Markov Models (HMMs)
and their many variants are often trained using `hard EM' \citep[][]{JuangRabiner1990,JordanEtAl1997,CohenSmith2010,AllahverdyanGalstyan2011}. TV-EM generalizations would in this case provide promising
future generalizations especially when noting that algorithms estimating multiple winning states are sometimes readily available \citep[e.g.][]{Foreman1992,HuangEtAl2012}. Again, TV-EM may also
serve to interpret earlier combinations of `hard EM' and standard EM for HMMs \citep[e.g.][]{AllahverdyanGalstyan2011,SpitkovskyEtAl2011} on the ground of a variational EM framework.

But also for long-standing standard tasks such as clustering, the here obtained results can be of direct theoretical and practical relevance.
By applying TV-EM to a special case of GMMs (isotropic and equally weighted Gaussians), \citet[][]{LuckeForster2019} have shown, for instance,
that the optimization of cluster centers decouples from the optimization of cluster variance for $\KKn$ with just one element (while the same is not true for $|\KKn|>1$ or for general GMMs).
The optimization of cluster centers is then observed to be equivalent to $k$-means. 
% 
%The equivalence is notably obtained without the requirement of taking the limit to zero cluster variances \citep[which is the standard textbook procedure to
%relate GMMs and $k$-means, e.g.,][]{MacKay2003,Barber2012}. %Furthermore, TV-EM for GMMs provides a free energy objective which is provably (and in this case strictly) increased by $k$-means.
%The objective bounds the GMM log-likelihood from below and is a function of the $k$-means objective, i.e., of the quantization error \citep[see][for details]{LuckeForster2017}. 
%Early speculations of this link to $k$-means can again be found at the end of \citep[][]{NealHinton1998} (but no results or proofs are given, and would have required variational
%distributions with `hard' zeros, and results such as Prop.\,1 to 5.

Finally, note that Prop.\,8 applies in general for any generative model, which implies that TV-EM generalizations of any previous `hard EM' approach are possible using TV-EM.
%and which (B)~suggests that free energy results as for $k$-means and GMMs \citep[][]{LuckeForster2017} may be obtained also for other generative models with discrete latents.

\paragraph{Relation to Sampling.} While standard variational EM procedures are typically regarded as being deterministic, TV-EM shares many properties with stochastic (i.e., sampling based) EM approaches. Like sampling approaches, TV-EM approximates probabilities (in our case posteriors) by a set of states in hidden space, and these states are then used to compute expectation values. Indeed, one option to realize concrete TV-EM optimizations is to suggest new states for $\KK$ by sampling from appropriate distributions (Sec.\,\ref{SecSampling}). TV-EM also shares with sampling that the accuracy of the approximation is only limited by computational demand. In the limit of infinite computational resources both sampling and TV-EM converge to EM with exact E-steps. The same can not be said about standard variational approaches like mean-field or Gaussian. TV-EM distinguishes itself from sampling, however, by being a variational approach that optimizes a free energy using variational distributions. While truncated approximations {\em can} make use of sampling as part of their optimization, sampling is just one option and other procedures like selection functions or other deterministic procedure (compare Sec.\,\ref{SecMixtureModels}) can be applied. Also the distributions that are used to suggest
states for $\KK$ are not limited to posterior distributions as also other (potentially easier to use) distributions can be sufficiently efficient in providing samples which increase the truncated free energy.
%\myvanish{Related to this point, issues like burn-in times or convergence times to an equilibrium distributions are not a direct concern within the TV-EM framework. A further difference is that the states in $\KK$ are variational parameters and as such should be memorized (at least partly) across EM iterations.}
%
While these are all points of difference, the fact that both TV-EM and sampling are approximating posteriors based on finite sets of hidden states, makes TV-EM the variational approximation that is most closely related to sampling. Also other combinations of variational approaches and sampling have been investigated (see discussion of `black box' approaches below) but none directly treats samples as variational parameters.
\paragraph{Autonomous and General Purpose (`Black Box') Inference and Learning.} Other than providing a general procedure to develop a learning algorithm for a specific generative model of interest, TV-EM is also of relevance for the field of autonomous machine learning, which recently attracted a lot of attention \citep[][]{RezendeMohamed2015,TranEtAl2015,RanganathEtAl2015,HernandezEtAl2016}. The goal of this field of study is to provide procedures that minimize expert intervention in the generation and application of learning and inference algorithms. Typically, user intervention is required for a 
number of steps in the process of developing a concrete learning algorithm for a given model.
Both standard variational approaches and standard sampling have to overcome different analytical and practical computational challenges. A factored variational EM approach, for instance, first has to choose a specific form for the variational distributions, and then requires a potentially significant effort to derive update equations for their variational parameters. Instead, TV-EM does not require additional analytical steps for variational E-step equations, expectation values are computed directly based on the variational states. Also in case of sampling, deriving, e.g., an efficient sampler requires additional and potentially highly non-trivial analytical work. On the other hand, the optimization of $\KK$ for TV-EM may require additional efforts. For previous truncated approximations, the construction of $\KK$ using preselection was model dependent (see Sec\,\ref{SecSampling}). However, given the novel results of this work, previous model specific constructions can be replaced by a general purpose optimization of $\KK$. If such an optimization is provided, then TV-E-steps are obtained which just use the joint probability of the generative model under consideration. For the M-step, a similar model independence can be achieved, e.g., by using automatic differentiation techniques, which are continuously further developed. The explicit form of TV-EM in Sec.\,\ref{SecExplicitForm} makes all the requirements of the algorithms very explicit and salient.

The formulation in terms of joint probabilities (\ref{EqnTVEMExplicit}) also relates TV-EM to research on unnormalized statistical models \citep[e.g.][and references therein]{GutmannHyvarinen2012}. %Unnormalized models as considered by \citet[][]{GutmannHyvarinen2012}, for instance, are trained very differently to TV-EM, however. 
The basic idea of the approach, e.g., by \citet[][]{GutmannHyvarinen2012}, is the use of classifiers to discriminate between observed and generated data, which is not used by TV-EM. Further differences are our focus on discrete latents, and our requirement for computationally tractable joint probabilities. This latter requirement was for classifier-based training 
dropped in later extensions, e.g., by \citet[][]{GutmannCorander2016}. On the other hand, classifier-based training \citep[also compare][]{GoodfellowEtAl2014} usually requires the definition of comparison metrics in observed space which is not used by TV-EM. % is not using such an additional . % TV-EM may offer  Similar such definitions (which may have strong influences on model training) are not required for TV-EM.

By considering one TV-EM iteration in its explicit form (\ref{EqnTVEMExplicit}), also note that TV-EM is very different, e.g., (A)~from \citet[][]{RezendeMohamed2015} who use normalizing flows and consider continuous latents, (B)~from \citet[][]{RanganathEtAl2015} who successively apply mean-field approach to realize latent dependencies, or (C)~from work by \citet[][]{TranEtAl2015} based on copulas. Further related studies are work by \citet[][]{SalimansEtAl2015} who use Markov Chain Monte Carlo (MCMC) samplers and treat the samples as auxiliary variables within a standard variational free energy framework as well as studies using stochastic variational inference \citep[][]{HoffmanEtAl2013,HoffmanBlei2015}, where auxiliary distributions for Markov chains are defined and used to approximate true posteriors. Moreover, 
\citet[][]{GuEtAl2015} use variational distributions as proposal distributions to realize flexible and efficient MCMC sampling.  
All these approaches are more indirect than the direct treatment of latent states as variational parameters as done by TV-EM. \citet[][]{SalimansEtAl2015}, for instance, also make a number of choices to define appropriate MCMC samplers (and they focus on models with continuous latents), \citet[][]{HoffmanBlei2015} do require sampling to estimate analytical intractabilities for their variational lower bound,
and \citet[][]{GuEtAl2015} use variational distributions as a means to improve approximations by of MCMC sampling. For TV-EM, sampling is one option to vary the variational parameters,
the procedure is by definition tractable for sufficiently small $\KKn$, and lower bounds are provably monotonically increased. On the other hand, TV-EM is constrained to discrete latents, and (as a novel approach) the performance for many concrete applications (which has been demonstrated for the other approaches discussed) remains to be explored.

In any case, the very active research on such and other very recent methods for a `black-box' optimization framework highlight (A)~the requirement for powerful and efficient approaches for advanced data models,
and (B)~the short-comings of conventional approximations. The use of collections of hidden states within variational approaches, e.g., as done by \citet[][]{SalimansEtAl2015,HoffmanBlei2015,GuEtAl2015} or with truncated approaches \citep[][]{LuckeEggert2010,SheltonEtAl2011,SheikhEtAl2014,SheikhLucke2016}, seems to be a promising general strategy in this respect. 
%
% ALSO ADD SIMULATOR BASED TRAINING, CONTRASTIVE DIVERGENCE, ETC   HoffmanEtAl2013, RezendeMohamed2015,   SalimansEtAl2015
%
% EXPLICIT FORM HIGHLIGHT THAT NO OTHER ADVANCED TRICKS HAVE TO BE USED - maybe mention that in explicit section
%

\paragraph{Outlook and Conclusion.} The theoretical framework of free energy optimization \citep[e.g][]{SaulEtAl1996,NealHinton1998,JordanEtAl1999} has been and is of exceptional significance
for the development of Machine Learning algorithms. In this work we extend the framework to include variational distributions with `hard' zeros but potentially many non-zero probabilities.
The distinugishing feature of the approach is that the sets of states with non-zero probabilities are treated as variational parameters.
%with exact zeros and with latent states as variational parameters. 
The theoretical results derived from these initial assumptions provided us with concise and easily applicable variational EM algorithms as well as concise and tractable forms of free energies.
The derived results, consequently, (A)~allow for developing novel algorithms, and (B)~allow for generalizations of the very established `hard EM' approaches.
%
%By combining truncated variational distributions with sampling approahce
%
Examples for the development of novel algorithms are combinations of variational EM with sampling or preselection methods (Sec.\,\ref{SecSampling}), or novel algorithms for
mixture models (Sec.\,\ref{SecMixtureModels}). Approaches that the here derived results embed into a variational free energy framework are relatively recent algorithms based
constrained likelihood optimizations by \citet[][]{HughesSudderth2016}, and earlier truncated approximations \citep[][]{LuckeSahani2008,LuckeEggert2010,HennigesEtAl2010,DaiLucke2014,SheikhEtAl2014}.
Furthermore, `hard EM' including partial hard E-steps is obtained as a special case of TV-EM. Still, Secs.\,\ref{SecSampling}, \ref{SecMixtureModels} and \ref{SecHardEM} are example applications of
the main theoretical results. Further applications may include future derivations of algorithms other than Alg.\,1 and 2. 
%The relation between k-means and Gaussian mixtures as studied by
%\citet[][]{LuckeForster2017} may serve as an example for such an application.
Examples for future applications of the here derived results may include generalizations of Viterbi training for
HMMs \citep[][]{JuangRabiner1990,JordanEtAl1997,CohenSmith2010,AllahverdyanGalstyan2011}
or applications to train deep networks \citep[][]{PoonDomingos2011,OordEtAl2014,PatelEtAl2016}. 
%
%More generally, future new learning algorithms can be based on any approach which efficiently addresses the very generally formulated E-step optimization problem (\ref{EqnTVEStep}).  Any flexible and generic solution developed will then instantly be of relevance for 'black box' approaches. As most visible when considering the explicit form of the here considered free energy optimization (Sec.\,\ref{SecExplicitForm}), the requirement for the generative models are (except of discrete latents) essentially reduced to the tractability of their joint probabilities. 

%\setcounter{page}{1}
\setcounter{subsection}{0}
\renewcommand{\thesubsection}{Appendix~\Alph{subsection}}

\section*{Appendix}

\subsection{\hspace{-2mm}: Free Energies for Non-Zero Variational Distributions}
\label{AppA}
Derivations of the standard results for free energy optimization are, e.g., given in \citep[see, e.g.,][]{SaulEtAl1996,Murphy2012,Barber2012}. In the notation used here,
first recall Jensen's inequality, which can for our purposes be denoted as follows:
Let $f$, $g$  and $q$ be real-valued functions such that $g(\sVec),f(\sVec),q(\sVec)\in\RRR$ and let $\Omega$ be the functions' domain. For all $\sVec\in\Omega$ let $q(\sVec)$ be a non-negative function that sums to one, i.e., $\sum_{\sVec\in\Omega}\,q(\sVec)=1$. Then for any {\em concave} function $f$ the
following inequality holds:
\begin{eqnarray}
  f\big(\sum_{\sVec}\,q(\sVec)\,g(\sVec)\big)\,\geq\, \sum_{\sVec}\,q(\sVec)\,f(g(\sVec))\,.
\label{EqnJensen}
\end{eqnarray}
Note that $\sum_{\sVec}$ in (\ref{EqnJensen}) again denotes a summation over all states $\sVec$. % In the following this set of all states $\sVec$ will, if required, be denoted by $\Omega$.
Any distribution $q(\sVec)$ on $\Omega$ satisfies the conditions on $q$ for Jensen's inequality. If we additionally demand $q(\sVec)$ to be {\em strictly} positive, i.e.\ $q(\sVec)>0$ for all $\sVec$,
we can apply Jensen's inequality to the data likelihood (\ref{EqnLikelihood}) because the logarithm is a concave function. We obtain:
\begin{eqnarray}
  \LL(\Theta) &=& \sum_{n=1}^{N}\ \log\big(\sum_{\sVec} p(\sVec,\yVecN\,|\,\Theta)\big)\ =\ \sum_{n=1}^{N}\ \log\big( \sum_{\sVec}\,\qn(\sVec)\,\frac{1}{\qn(\sVec)}\,p(\sVec,\yVecN\,|\,\Theta)\big)\phantom{\int^f_g}\nonumber\\
                     \,&\geq&\,\,  \sum_{n=1}^{N}\ \sum_{\sVec}\,\qn(\sVec)\, \log\big( \frac{1}{\qn(\sVec)}\,p(\sVec,\yVecN\,|\,\Theta) \big)\phantom{\int^f_g}\nonumber\\
                     \,&=&\,\, \sum_{n=1}^{N}\Big( \sum_{\sVec}\,\qn(\sVec)\,\log\big(p(\sVec,\yVecN\,|\,\Theta)\big)\,-\,\sum_{\sVec}\,\qn(\sVec)\,\log\big(\qn(\sVec)\big)\Big)\,.\phantom{\int^f_g}
%
  %                   \,&=&\, \E{\log\big(p(\sVec,\yVecN)\big) \,-\, \SSS{q(\sVec;\Lambda)}\,&=:&\, \FFvar (\beta,\Theta;\Lambda) \phantom{\int^f_g}\,.\phantom{\int^f_g}
%
\label{EqnFreeEnergyDeri}
\end{eqnarray}
%
%where $\sum_{\sVec}$ again denotes the sum over all states of $\sVec$.
The crucial novel entity that emerges in this derivation
is a lower bound of the likelihood which is termed the (variational) {\em free energy} \citep[compare][]{SaulJordan1996,NealHinton1998,JordanEtAl1999,MacKay2003}:
%
%\begin{multline}
\begin{eqnarray}
\label{EqnFreeEnergy}
  \LL(\Theta) \,\geq\, \FF(q,\Theta)\,:=\,
    \sum_{n=1}^{N} \Big(
      \sum_{\sVec}\ \qn(\sVec)\  
       \log\big( p(\sVec,\yVecN\,|\,\Theta)\big)
      \Big) 
      + H(q)\,,
\end{eqnarray}
%\end{multline}
%
where $H(q) = - \sum_n\sum_{\sVec} \qn(\sVec) \log(\qn(\sVec))$ is a
function (the Shannon entropy) that depends on the distribution
$q(\sVec)=q^{(1:N)}(\sVec)=(q^{(1)}(\sVec),\ldots,q^{(N)}(\sVec))$ but not on the model parameters~$\Theta$.
Given a data point $\yVecN$, the distribution $q^{(n)}(\sVec)$ is called {\em variational
  distribution}. Also the collection of all distributions $q(\sVec)$ , with $q(\sVec)=(q^{(1)}(\sVec),\ldots,q^{(N)}(\sVec))$, is
referred to as variational distribution. %Note that we require $q^{(n)}(\sVec)>0$ for all $\sVec$ and all $n$.
The name `free energy' was inherited from statistical physics where 
approximations to the Helmholtz free energy of a physical system take a very similar mathematical form \citep[see, e.g.,][]{MacKay2003}.
% More precisely, while the analog to the Helmholtz free energy in physics is the data log-likelihood, the analog to variational free energies
%in physics is the variational free energy in (\ref{EqnFreeEnergy}).
%
%\changedOther{In Machine Learning, the sum over all states of $\sVec$ in \refp{EqnFreeEnergy} becomes an integral
%if the values of $\sVec$ are continuous.} 

%In the EM scheme $\FF(q,\Theta)$ is maximized alternately with respect to $q$ in
%the E-step (while $\Theta$ is kept fixed) and with respect to $\Theta$
%in the M-step (while $q$ is kept fixed).  Using EM it can thus be shown
%that the EM iterations monotonically increase the likelihood. In practical
%applications EM is found to increase the likelihood to (at least
%local) likelihood maxima.

\subsection{\hspace{-2mm}: The Free Energy for Variational Distributions with `hard' Zeros}
\label{AppB}
We provide the proofs for Props.\,1 and 2 which formally extend the basic variational EM results to include variational distributions with exact zeros.

\ \\
{\bf Proof of Proposition 1}\\*
First let us (as was also done for Eqn.\,\ref{EqnPropOne}) drop the notation of variational parameters to increase readability. Then observe that the
standard derivation (\ref{EqnFreeEnergyDeri}) shows that the proposition is true if $\qn(\sVec)>0$ for all $\sVec\in\Omega$ and all $n$.
To show that (\ref{EqnPropOne}) is true for any distribution, consider the case of variational distributions $\qn(\sVec)$
for which $\qn(\sVec)>0$ is true only in proper subsets $\KKn$ of the state space $\Omega$, and
equal to zero for all $\sVec\not\in\KKn$. For a given data point $n$ a distribution is either $\qn(\sVec)>0$ for all $\sVec$ or there exists a proper
subset $\KKn$. In the latter case, it is trivial to define such a $\KKn$, and as $\qn(\sVec)$ is a distribution (i.e., sums to one), this implies that $\KKn$ is not empty.
For the main part of the proof we will now assume that for all $n$ there exists a proper subset $\KKn$ of $\Omega$.
A general $q$ may consist of distributions $\qn(\sVec)$ that are strictly positive for some $n$ and have proper subsets $\KKn$
for the other $n$. However, addressing these mixed cases will turn out to be straight-forward, and we will come back to this general case at the end of the proof.
%
% size $|\KK|<2^H$ (assuming binary states of $H$ variablesw). 

Given a distribution $\qn(\sVec)$ with set $\KKn$, let us define an auxiliary function $\qnt(\sVec)$ as follows:
\begin{equation}
\qnt(\sVec) = \left\{
%
%$$
\begin{array}{ll}
\qn(\sVec)\,-\,\epsn^-  &  \mbox{for all}\ \sVec\in\KKn\\
\qn(\sVec)\,+\,\epsn^+  &  \mbox{for all}\ \sVec\not\in\KKn
\label{EqnQt}
\end{array}
%$$
\right.
\end{equation}
Let $\epsn^-$ be greater zero but smaller than any value of $\qn(\sVec)$ in $\KKn$, i.e.,
\begin{eqnarray}
0 \,<\,\epsn^- \,<\, \min_{\sVec\in\KKn}\{\qn(\sVec)\}\phantom{iiiii}\mbox{and let}\phantom{iiiii}\epsn^+ \,:=\, \frac{|\KKn|}{|\Omega|\,-\,|\KKn|}\,\epsn^-\,,
\label{EqnEpsMinus}
\label{EqnEpsPlus}
\end{eqnarray}
where $|\KKn|$ is the number of states in $\KKn$ and where $|\Omega|$ is the number of all possible states. %, e.g., $2^H$ for $H$ binary variables.
As we have defined $\KKn$ to contain only values $\qn(\sVec)>0$, the minimum in $\KKn$ is greater zero, i.e.,
we can always find an $\epsn^-$ satisfying (\ref{EqnEpsMinus}). Consequently, $\epsn^+$ is also greater zero and finite 
as we demanded the $\KKn$ to be proper subsets of $\Omega$ ($|\KKn|<|\Omega|$).

With definitions (\ref{EqnEpsMinus}) observe that $\qnt(\sVec)>0$ for all $\sVec$ and that $\sum_{\sVec}\qt(\sVec)=1$, i.e., $\qnt(\sVec)$ is a distribution on $\Omega$
which satisfies (other than $\qn(\sVec)$) the requirement for the original derivation of the variational free energy (\ref{EqnFreeEnergyDeri}). We can therefore use the
free energy definition for $\qt(\sVec)=\big(\qt^{(1)}(\sVec),\ldots,\qt^{(N)}(\sVec)\big)$ and then consider the limit to small $\epsn^-$.
For this, we insert the definition of $\qnt(\sVec)$ into the free energy and find:

\newcommand{\spacc}{\hspace{-8mm}}
\footnotesize
$$
\begin{array}{rcll}
\disS
&&\spacc{}\LL (\Theta) \,\geq\,\disS \FF(\qt,\Theta)\,\phantom{\int^f_g}\nonumber\\
&&\spacc{}=\disS\,\sum_n\Big( \sum_{\sVec}\,\qnt(\sVec)\,\log\big(p(\sVec,\yVecN\,|\,\Theta)\big)\,-\,\sum_{\sVec}\,\qnt(\sVec)\,\log\big(\qnt(\sVec)\big) \Big)\,\phantom{\int^f_g}\nonumber\\
%
%&&\spacc{}=\disS\,\sum_n\Big( \sum_{\sVec}\,\qnt(\sVec)\,\,\log\big(p(\sVec,\yVecN\,|\,\Theta)\big)\,-\,\sum_{\sVec}\,\qnt(\sVec)\,\log\big(\qnt(\sVec)\big) \Big)\phantom{\int^f_g} \\
%
%&&\spacc{}=\disS\,\\sum_n\Big( sum_{\sVec\in\KKn}\,\qnt(\sVec)\,\,\log\big(p(\sVec,\yVecN\,|\,\Theta)\big)\,-\,\sum_{\sVec\in\KKn}\,\qnt(\sVec)\,\log\big(\qnt(\sVec)\big)\phantom{\int^f_g} \\
%&&\spacc{}\disS\,  \phantom{=}+\,\sum_{\sVec\not\in\KKn}\,\qnt(\sVec)\,\,\log\big(p(\sVec,\yVecN\,|\,\Theta)\big)\,-\,\sum_{\sVec\not\in\KKn}\,\qnt(\sVec)\,\log\big(\qnt(\sVec)\big) \Big)\phantom{\int^f_g} \\
%
&&\spacc{}=\disS\,\sum_n\Big( \sum_{\sVec\in\KKn}\,\big( \qn(\sVec)-\epsn^-\big) \,\log\big(p(\sVec,\yVecN\,|\,\Theta)\big)\,-\,\sum_{\sVec\in\KKn}\,\big( \qn(\sVec)-\epsn^-\big)\,\log\big(\qn(\sVec)-\epsn^-\big)\phantom{\int^f_g} \\
&&\spacc{}\disS\,  \phantom{=}+\,\sum_{\sVec\not\in\KKn}\,\big( \qn(\sVec)+\epsn^+ \big)\,\log\big(p(\sVec,\yVecN\,|\,\Theta)\big)\,-\,\sum_{\sVec\not\in\KKn}\,\big( \qn(\sVec)+\epsn^+ \big)\,\log\big(\qn(\sVec)+\epsn^+\big) \Big)\phantom{\int^f_g} \\
&&\spacc{}=\disS\,\sum_n\Big( \sum_{\sVec}\,\qn(\sVec)\,\log\big(p(\sVec,\yVecN\,|\,\Theta)\big) \,-\, \epsn^-\,\,\sum_{\sVec\in\KKn}\,\log\big(p(\sVec,\yVecN\,|\,\Theta)\big) + \epsn^+\,\,\sum_{\sVec\not\in\KKn}\,\log\big(p(\sVec,\yVecN\,|\,\Theta)\big)\phantom{\int^f_g}\\
&&\spacc{}\disS\,  \phantom{=}-\,\sum_{\sVec\in\KKn}\,\big( \qn(\sVec)-\epsn^-\big)\,\log\big(\qn(\sVec)-\epsn^-\big) -\epsn^+ \,\log(\epsn^+)\,\sum_{\sVec\not\in\KKn}\,1\phantom{\int^f_g} \Big) \\
%
%
%\end{array}
%$$
%
%$$
%\begin{array}{rcll}
%
%\disS
%
&&\spacc{}=\disS\,\sum_n\Big( \sum_{\sVec}\qn(\sVec)\,\log\big(p(\sVec,\yVecN\,|\,\Theta)\big) - \epsn^-\sum_{\sVec\in\KKn}\,\log\big(p(\sVec,\yVecN\,|\,\Theta)\big) + \epsn^-\,\frac{|\KKn|}{|\Omega|-|\KKn|}\sum_{\sVec\not\in\KKn}\log\big(p(\sVec,\yVecN\,|\,\Theta)\big)\phantom{\int^f_g} \\
&&\spacc{}\disS\,  \phantom{=}-\, \sum_{\sVec\in\KKn}\,\big( \qn(\sVec)-\epsn^-\big)\,\log\big(\qn(\sVec)-\epsn^-\big) - \epsn^- \,|\KKn|\,\log\big(\frac{|\KKn|}{|\Omega|-|\KKn|}\big) - |\KKn|\,\epsn^- \,\log(\epsn^-) \Big)   \phantom{\int^f_g} \\
&&\spacc{}=\disS\,\sum_n\Big( \sum_{\sVec}\,\qn(\sVec)\,\log\big(p(\sVec,\yVecN\,|\,\Theta)\big) - \sum_{\sVec\in\KKn}\,\big( \qn(\sVec)-\epsn^-\big)\,\log\big( \qn(\sVec)-\epsn^- \big) - |\KKn|\,\epsn^- \,\log(\epsn^-) \phantom{\int^f_g} \\ 
&&\spacc{}\disS\,  \phantom{=}-\, \epsn^-\sum_{\sVec\in\KKn}\,\log\big(p(\sVec,\yVecN\,|\,\Theta)\big) + \epsn^-\frac{|\KKn|}{|\Omega|-|\KKn|}\,\,\sum_{\sVec\not\in\KKn}\,\log\big(p(\sVec,\yVecN\,|\,\Theta)\big) - \epsn^- \,|\KKn|\,\log\big(\frac{|\KKn|}{|\Omega|-|\KKn|}\big) \Big) \phantom{\int^f_g}
\end{array}
$$
\normalsize
%
%                     \,&=&\disS\,          \phantom{+\,}\,\sum_{\sVec\in\KKn}\,\big( \qn(\sVec)-\epsn^-\big) \,\log\big(p(\sVec,\yVecN\,|\,\Theta)\big)+\sum_{\sVec\in\KKn}\,\big( \qn(\sVec)\,-\,\epsn^-\big)\,\log\big(\qn(\sVec)\,-\,\epsn^-\big)\phantom{\int^f_g} \\
%                     \,&\phantom{=}&\disS\,          +\,\,\sum_{\sVec\not\in\KKn}\,\big( \qn(\sVec)+\epsn^+ \big)\,\log\big(p(\sVec,\yVecN\,|\,\Theta)\big)+\sum_{\sVec\not\in\KKn}\,\big( \qn(\sVec)+\epsn^+ \big)\,\log\big(\qn(\sVec)+\eps^+\big)\phantom{\int^f_g} \\
%
This expression applies for any $\epsn^-$ satisfying (\ref{EqnEpsMinus}) and therefore also if we replace:
%
%\footnotesize
\begin{eqnarray}
\epsn^- \,=\, \eps \,<\, \min_{n^\prime=1,\ldots,N}\{\eps_{n^\prime}^-\},
%
%\label{EqnConvSecond}
\end{eqnarray}
\normalsize
such that we obtain:
\footnotesize
\begin{equation}
\begin{array}{rcll}
\disS
&&\spacc{}\hspace{-4mm} \FF (\qt,\Theta)=\disS\,\sum_n\Big( \sum_{\sVec}\,\qn(\sVec)\,\log\big(p(\sVec,\yVecN\,|\,\Theta)\big)\,-\, \sum_{\sVec\in\KKn}\,\big( \qn(\sVec)-\eps\big)\,\log\big( \qn(\sVec)-\eps \big) - |\KKn|\,\eps \,\log(\eps) \hspace{-9mm}\phantom{\int^f_g} \\ 
&&\spacc{}\hspace{-4mm} \disS\,  \phantom{=}-\, \eps\sum_{\sVec\in\KKn}\,\log\big(p(\sVec,\yVecN\,|\,\Theta)\big) + \eps\,\frac{|\KKn|}{|\Omega|-|\KKn|}\,\,\sum_{\sVec\not\in\KKn}\,\log\big(p(\sVec,\yVecN\,|\,\Theta)\big) - \eps \,|\KKn|\,\log\big(\frac{|\KKn|}{|\Omega|-|\KKn|}\big) \Big) \hspace{-9mm}\phantom{\int^f_g}
\end{array}
\label{EqnFFInter}
\end{equation}
\normalsize
Let us now consider infinitesimally small $\eps>0$, i.e., let us consider the limit when $\eps\rightarrow{}0$.
%For instance, we can choose a sequence $\eps^{(k)}=\frac{1}{k}$ starting at sufficiently large $k$
%as there is always a $K$ such that (\ref{EqnConvSecond}) is satisfied for all $k>K$.
First, observe that in this case all summands of the last line of $\FF (\qt,\Theta)$ in (\ref{EqnFFInter}) trivially converge to zero. Then observe that the second summand of $\FF (\qt,\Theta)$ converges to
%
%\footnotesize
\begin{eqnarray}
\lim_{\eps\rightarrow{}0} \Big( \sum_{\sVec\in\KKn}\,\big( \qn(\sVec)\,-\,\eps\big)\,\log\big( \qn(\sVec)\,-\,\eps \big) \Big)
\,=\, \sum_{\sVec\in\KKn}\,\qn(\sVec)\,\log\big( \qn(\sVec) \big)
%
%\label{EqnConvSecond}
\end{eqnarray}
\normalsize
as $\qn(\sVec)$ is finite and greater zero for all $\sVec\in\KKn$. Finally, the third summand in (\ref{EqnFFInter}) can be observed to converge (following l'H\^opital) to zero:
%
%\footnotesize
\begin{eqnarray}
\lim_{\eps\rightarrow{}0} \big( \eps \,\log(\eps) \big)
\,=\, \lim_{\eps\rightarrow{}0} \big( \frac{\log(\eps)}{\frac{1}{\eps}} \big)
\,=\, \lim_{\eps\rightarrow{}0} \big( -\frac{\frac{1}{\eps}}{\frac{1}{\eps^2}} \big)
\,=\, -\,\lim_{\eps\rightarrow{}0} \big( \eps \big) \,=\,0\,.
\label{EqnConvSecond}
\end{eqnarray}
\normalsize
Thus, the limit $\eps\rightarrow{}0$ exists and is given by:
\footnotesize
\begin{equation}
\disS\hspace{1mm}
\FF (q,\Theta)\,=\,\lim_{\eps\rightarrow{}0} \FF (\qt,\Theta) \,=\,\disS\,\sum_n \sum_{\sVec}\,\qn(\sVec)\,\log\big(p(\sVec,\yVecN\,|\,\Theta)\big)-\sum_n\sum_{\sVec}\, \qn(\sVec)\,\log\big( \qn(\sVec) \big), \hspace{-5mm}\phantom{\int^f_g} \normalsize
\label{EqnFFLimitRes}
\end{equation}
\normalsize
where we have used (because of Eqn.\ \ref{EqnConvSecond}) the convention $\qn(\sVec)\,\log\big( \qn(\sVec) \big) = 0$ for all $\qn(\sVec)=0$.
We will use this convention (which is also commonly used for the KL-divergence) throughout the paper.

Eqn.\,\ref{EqnFFLimitRes} shows that Proposition\,1 holds for any $q$ with distributions $\qn(\sVec)$ with proper subsets $\KKn$ of $\Omega$.
To finally show that Proposition\,1 holds for general $q$ (with discrete $\sVec)$, consider the mixed case that $q$ contains strictly positive distributions $\qn$ for some $n$ and
distributions $\qn$ with exact zeros for the other $n$'s. Let us define the set $J\subseteq\{1,\ldots,N\}$ to contain those $n$ with $\qn$ being strictly positive, and the complement $\JBar\subseteq\{1,\ldots,N\}$
to contain those $n$ with $\qn$ that are not strictly positive. Using $J$ and $\JBar$ we then define an auxiliary function $\qt$ by using $\qnt(\sVec)$ of Eqn.\,\ref{EqnQt} only
for $n\in\JBar$ and by setting $\qnt(\sVec)=\qn(\sVec)$ for all $\sVec$ for all $n\in{}J$.
The functions $\qnt$ are then again strictly positive distributions on $\Omega$ for all $n$
and we can again apply the standard result \refp{EqnFreeEnergyDeri}: \\
\footnotesize
$$
\begin{array}{rcll}
\disS
\spacc{}\LL (\Theta) \,\geq\,\disS \FF(\qt,\Theta)\phantom{ii}
&&\spacc{}=\disS\,\phantom{iiiii}\sum_{n}\Big( \sum_{\sVec}\,\qnt(\sVec)\,\log\big(p(\sVec,\yVecN\,|\,\Theta)\big)\,-\,\sum_{\sVec}\,\qnt(\sVec)\,\log\big(\qnt(\sVec)\big) \Big)\,\phantom{\int^f_g}\nonumber\\
&&\spacc{}=\disS\phantom{\phantom{ii}+\,}\sum_{n\in{}J}\Big( \sum_{\sVec}\,\qn(\sVec)\,\log\big(p(\sVec,\yVecN\,|\,\Theta)\big)\,-\,\sum_{\sVec}\,\qn(\sVec)\,\log\big(\qn(\sVec)\big) \Big)\,\phantom{\int^f_g}\nonumber\\
&&\spacc{}\phantom{iiiiii}+\disS\,\sum_{n\in\JBar}\Big( \sum_{\sVec}\,\qnt(\sVec)\,\log\big(p(\sVec,\yVecN\,|\,\Theta)\big)\,-\,\sum_{\sVec}\,\qnt(\sVec)\,\log\big(\qnt(\sVec)\big) \Big)\,\phantom{\int^f_g}\nonumber\\
\end{array}
$$\vspace{1mm}
\normalsize

\noindent{}Now we apply the same arguments as above but only to the sum over all $n\in{}\JBar$, for which everything remains as for the derivation above.
Eqn.\,\ref{EqnFFLimitRes} therefore applies if we only consider sums over $\JBar$.
Thus in the limit $\eps\rightarrow{}0$ we obtain in the mixed case:\vspace{1mm}

\footnotesize
\begin{eqnarray}
\disS
\FF (q,\Theta) &=& \lim_{\eps\rightarrow{}0}\big( \FF (\qt,\Theta) \big) \nonumber\\
&=&\disS\sum_{n\in{}J}\Big( \sum_{\sVec}\,\qn(\sVec)\,\log\big(p(\sVec,\yVecN\,|\,\Theta)\big)\,-\,\sum_{\sVec}\,\qn(\sVec)\,\log\big(\qn(\sVec)\big) \Big)\,\phantom{\int^f_g}\nonumber\\
&&\spacc{}\phantom{iiiiii}+\disS\,\lim_{\eps\rightarrow{}0}\Big( \sum_{n\in\JBar}\Big( \sum_{\sVec}\,\qnt(\sVec)\,\log\big(p(\sVec,\yVecN\,|\,\Theta)\big)\,-\,\sum_{\sVec}\,\qnt(\sVec)\,\log\big(\qnt(\sVec)\big) \Big) \Big) \,\phantom{\int^f_g}\nonumber\\
%
%&=&\disS\sum_{n\in{}J}\Big( \sum_{\sVec}\,\qn(\sVec)\,\log\big(p(\sVec,\yVecN\,|\,\Theta)\big)\,-\,\sum_{\sVec}\,\qn(\sVec)\,\log\big(\qn(\sVec)\big) \Big)\,\phantom{\int^f_g}\nonumber\\
%
%&&\spacc{}\phantom{iiiiii}+\disS\,\sum_{n\in\JBar}\Big( \sum_{\sVec}\,\qn(\sVec)\,\log\big(p(\sVec,\yVecN\,|\,\Theta)\big)\,-\,\sum_{\sVec}\,\qn(\sVec)\,\log\big(\qn(\sVec)\big) \Big)\,\phantom{\int^f_g}\nonumber\\
%
&=&\disS\sum_{n}\Big( \sum_{\sVec}\,\qn(\sVec)\,\log\big(p(\sVec,\yVecN\,|\,\Theta)\big)\,-\,\sum_{\sVec}\,\qn(\sVec)\,\log\big(\qn(\sVec)\big) \Big)\,\phantom{\int^f_g}\nonumber
%
%&&\spacc{}\phantom{iiiiii}+\disS\,\sum_{n\in\JBar}\Big( \sum_{\sVec}\,\qn(\sVec)\,\log\big(p(\sVec,\yVecN\,|\,\Theta)\big)\,-\,\sum_{\sVec}\,\qn(\sVec)\,\log\big(\qn(\sVec)\big) \Big)\,\phantom{\int^f_g}\nonumber\\
%
\label{EqnFFLimitResMixed}
\end{eqnarray}
\normalsize
which finally shows that Proposition\ 1 holds for any $q$ with any distributions $\qn$ on $\Omega$.\nopagebreak\\
\BOX\\
\ \\
In principle, a function can be a lower bound for any finite value of (in this case) $\eps$ but the limit point can cease to maintain this property.
We thus formally show that the free energy remains a lower bound also in the limit, which is possible using the 
KL-divergence result (\ref{EqnDKLStandard}) and a similar approach as for the proof above.

\noindent{}{\bf Proof of Proposition 2}\\*
Given distributions $\qn(\sVec)$ let us consider the same auxiliary distributions $\qnt(\sVec)$ as for the proof of Prop,\,1 (see Eqn.\,\ref{EqnQt}).
As $\qnt(\sVec)$ are strictly positive distributions, Eqn.\,\ref{EqnDKLStandard} applies:% \citep[e.g.][]{Bishop2006,MacKay2003,Barber2012}.
%
%it applies according to the standard derivation of Eqn.\,\ref{EqnDKLStandard} \citep[e.g.][]{Bishop2006,MacKay2003,Barber2012}.
%(see Eqn.\,\ref{EqnDKLStandard} and Appendix):
%
\begin{eqnarray}
%\label{EqnDKLProofTwo}
%
\LL(\Theta)-\FF(\qt,\Theta)\ =\ \sum_{n}\DKL{\qnt(\sVec)}{p(\sVec\,|\,\yVecN,\Theta)}\ \geq\ 0,
\end{eqnarray}
for any collection of $\epsn^-$ satisfying (\ref{EqnEpsMinus}). If we now consider a sequence $\eps_k=1/k$, we know that for any $n$ there exists
a finite $K$ such that for all $k>K$ applies that $\epsn^-=\eps_k=1/k$ fulfills condition (\ref{EqnEpsMinus}). If we now set $\epsn^-=\eps_k=1/k$
for all $n$, we know because of finitely many data points $n$ that there also exists a finite $K$ such that condition (\ref{EqnEpsMinus}) is
fulfilled for all $k>K$. If we now define a sequence of distributions $\qt_k(\sVec)=(\qt^{(1)}_k,\ldots,\qt^{(N)}_k)$ by choosing $\epsn^-=\eps_k=1/k$ for all $n$,
we know that for all $k>K$ applies:
\begin{eqnarray}
%\label{EqnDKLProofTwo}
%
D_k\ =\ \LL(\Theta)-\FF(\qt_k,\Theta)\ =\ \sum_{n}\DKL{\qnt_k(\sVec)}{p(\sVec\,|\,\yVecN,\Theta)}\ \geq\ 0.
\end{eqnarray}
The sequence $D_k$ is hence a sequence in the interval $[0,\infty)$. As the limit $\lim_{k\rightarrow\infty}\FF(\qt_k,\Theta)$ is finite,
$D_k$ converges to a finite value within $[0,\infty)$ (which is left-closed). If $q$ contains some strictly positive $\qn$, we only use
$\epsn^-=\eps_k=1/k$ for all $n\in\JBar$. Finally, by using Proposition 1, the limit $\lim_{k\rightarrow\infty}D_k$ is given
by:
\footnotesize
$$
\begin{array}{rcll}
\disS
&&\hspace{-8mm} 0\,\leq\,\disS\lim_{k\rightarrow\infty}D_k \,=\,\disS \LL(\Theta)-\lim_{k\rightarrow\infty}\FF(\qt_k,\Theta)\,\phantom{\int^f_g}\nonumber\\
&&\hspace{-6mm} \disS \,=\,\LL(\Theta)\,-\,\FF(q,\Theta) \phantom{\int^f_g}\nonumber\\
&&\hspace{-6mm} \disS \,=\,\LL(\Theta)\,-\,\sum_n \sum_{\sVec}\,\qn(\sVec)\,\log\big(p(\sVec,\yVecN\,|\,\Theta)\big)\,+\, \sum_n\sum_{\sVec}\, \qn(\sVec)\,\log\big( \qn(\sVec) \big) \phantom{\int^f_g}\nonumber\\
&&\hspace{-6mm} \disS \,=\,\LL(\Theta) -\sum_n \underbrace{\sum_{\sVec}\,\qn(\sVec)}_{1}\,\log\big(p(\yVecN\,|\,\Theta)\big)-\sum_n \sum_{\sVec}\,\qn(\sVec)\,\log\big(p(\sVec\,|\,\yVec,\Theta)\big)
+ \sum_n\sum_{\sVec}\, \qn(\sVec)\,\log\big( \qn(\sVec) \big) \phantom{\int^f_g}\nonumber\\
&&\hspace{-6mm} \disS \,=\,-\sum_n \Big( \sum_{\sVec}\,\qn(\sVec)\,\log\big(p(\sVec\,|\,\yVec,\Theta)\big) \,-\, \sum_{\sVec}\, \qn(\sVec)\,\log\big( \qn(\sVec) \big) \Big) \phantom{\int^f_g}
\hspace{-5mm}\,=\,\sum_n \DKL{\qn(\sVec)}{p(\sVec\,|\,\yVecN,\Theta)}\,,\phantom{\int^f_g}
\end{array}
$$
\normalsize
where the last part follows the lines of the standard derivation for the difference $\LL(\Theta)\,-\,\FF(q,\Theta)$. 
Note that we again used the 
convention $\qn(\sVec)\,\log\big( \qn(\sVec) \big) = 0$ for all $\qn(\sVec)=0$.\nopagebreak\\
\BOX

\subsection{\hspace{-2mm}: Properties of `hard EM' Free Energies}
\label{AppC}
The free energy objective \refp{EqnFreeEnergyHard} is trivially related to objective functions stated in previous work \citep[e.g.][]{JuangRabiner1990,CeleuxGovaert1992,CohenSmith2010,JordanEtAl1997}
and the properties of the free energy we give below can also be derived from such objectives directly. Here, we will derive them from the
rigorous derivations required for general truncated distributions including the formulation using partial E- and M-steps (Sec.\,\ref{SecPartialTVEM}).

Based on the results for TV-EM, the following applies: The free energy \refp{EqnFreeEnergyHard} for the `hard EM' algorithm is monotonically increased by each `hard EM' step (Alg.\,\ref{AlgHardEM}). Furthermore, it applies
that the free energy increases if
\begin{eqnarray}
\label{EqnHardEMPartial}
p(\sVecN\,|\,\yVecN,\ThetaOld)>p(\sVecN_{\mathrm{old}}\,|\,\yVecN,\ThetaOld),
\end{eqnarray}
where $\sVecN_{\mathrm{old}}$ are the states found in the previous `hard' E-step. Finally, after each `hard EM' iteration or partial `hard EM' iteration,
the difference between log-likelihood (\ref{EqnLikelihood}) and free energy (\ref{EqnFreeEnergyHard}) is given by:
\begin{eqnarray}
\label{EqnDiffHardEM}
\LL(\Theta) - \FF(\sVec^{(1:N)},\Theta) = - \sum_{n=1}^N \log\big(p(\sVecN\,|\,\yVecN,\Theta)\big).
\end{eqnarray}
%
%Formal proofs of all these statements are given in Appendix ... .
%
%As `hard EM' is widely applied, similar such relations can also be derived directly without considering `hard EM' as a special case of TV-EM. The objective function
%formulations mentioned above \citep[e.g.][]{JuangRabiner1990,CeleuxGovaert1992,CohenSmith2010} contain such relations at least
%for specific generative models.  
%
%
These properties of the `hard EM' free energy follow from results for TV-EM as follows:

According to Prop.\,8, `hard' EM is equivalent to TV-EM (with sets $\KKn=\{\sVecN\}$). Consequently,
Props.\,1 to 5 are applicable to `hard EM', including a guaranteed monotonic increase of the simplified truncated free energy (\ref{EqnTruncatedF}).
For $\KKn=\{\sVecN\}$ the truncated free energy $\FF(\KK,\Theta)$ is given by (\ref{EqnFreeEnergyHard}), where we replaced $\KK=(\{\sVec^{(1)}\},\ldots,\{\sVec^{(N)}\})$ by $\sVec^{(1:N)}$, which
proves the first claim.

According to the results for partial TV-EM (Sec.\,\ref{SecPartialTVEM}), the free energy (\ref{EqnFreeEnergyHard}) also monotonically increases for
a {\em partial} TV-E-step (Optimization 1 of Eqns.\,\ref{EqnTVEMOptStepsParialFinal}). For $\KKn=\{\sVecN\}$, the free energy (\ref{EqnFreeEnergyHard}) is
monotonically increased if for all $n$ applies $p(\sVecN,\yVecN\,|\,\ThetaOld)\geq{}p(\sVecN_{\mathrm{old}},\yVecN\,|\,\ThetaOld)$ (compare Prop.\,6). As the data points are
constant, this condition is equivalent to condition (\ref{EqnHardEMPartial}) which proves the claim, i.e., it is sufficient
to monotonically increase $p(\sVec\,|\,\yVecN,\ThetaOld)$ for all (or just some) $n$ given the current parameters $\ThetaOld$,
and starting from the previous MAP states $\sVecN_{\mathrm{old}}$.

According to Props.\,1 to 3, $\FF(\sVec^{(1:N)},\Theta)$ is, as a special case of $\FF(\sVec^{(1:N)},\ThetaOld,\Theta)$, a lower bound of the log-likelihood.
Furthermore, the difference $\LL(\Theta) - \FF(\sVec^{(1:N)},\Theta)$ is given by the KL-divergence $\sum_n\DKL{\qn(\sVec;\KK,\Theta)}{p(\sVec\,|\,\yVecN,\Theta)}$ (Corollary~1).
As the truncated variational distributions are here given by $\qn(\sVec;\KKnew,\ThetaOld)=\delta(\sVec=\sVecN)$, we obtain (\ref{EqnDiffHardEM}). 
The relation between $\LL(\Theta) - \FF(\sVec^{(1:N)},\Theta)$ and the KL-divergence also applies for partial TV-EM and consequently for partial `hard EM' as a special case.\\
\BOX\\
%
%
%
%
%\bibliography{../cnml-all}
%
%
%\bibliography{../../ml_bibs/cnml-all}
%\bibliographystyle{apalike}
%
%
%\bibliography{../../ml_bibs/cnml-all}
%\bibliographystyle{apalike}
%
%

%
\end{document}